\documentclass[acmsmall, nonacm]{acmart}

\usepackage{makecell}
\usepackage{multirow}
\usepackage[table,dvipsnames]{xcolor}
\definecolor{lightblue}{rgb}{0.85, 1.0, 1.0}

\usepackage{footmisc}
\usepackage{tabularx}
\usepackage{pifont}
\usepackage[ruled,vlined]{algorithm2e}

\AtBeginDocument{
  }

\begin{document}

\title[A Study on Urdu Katib Handwritten Dataset Generation and CRNN-Based Baseline Evaluation]{Urdu Katib Handwritten Dataset: A Historical Document Dataset for Offline Urdu Handwritten Text Recognition with CRNN-Based Baseline Evaluation}

\author{Ramza Basharat}
\email{ramzabasharat19@gmail.com}
\orcid{0009-0004-6750-1069}
\author{Muhammad Usman Ali}
\email{m.usmanali@uog.edu.pk}
\orcid{0000-0002-4470-8065}
\affiliation{
  \institution{Department of Computer Science, University of Gujrat}
  \city{Gujrat}
  \state{Punjab}
  \country{Pakistan}
}

\renewcommand{\shortauthors}{Ramza Basharat et al.}

\begin{abstract}
Automatic Handwritten Text Recognition (HTR) is inherently a challenging task, and its complexity is further increased when dealing with cursive scripts. Although significant efforts have been made on various cursive scripts, research regarding Urdu Handwritten Text Recognition (UHTR) has been relatively limited. This lag of research is primarily due to the unique challenges posed by its script, and the scarcity and unavailability of benchmark datasets. Therefore, to advance research in UHTR, this study presents a specialized real dataset called the \textit{Urdu Katib Handwritten Dataset (UKHD)}. To the best of our knowledge, this is the first offline Urdu handwritten text lines dataset specifically curated from the materials written by \textit{Katibs} in historical times. It encompasses a diverse range of flat nib writing variations in the Nastalique calligraphic style. Additionally, the effectiveness of different CRNN-based hybrid models has been evaluated to identify the optimal architecture for \textit{Urdu Katib Handwriting Recognition (UKHR)}. Among the analyzed models, the CNN-BGRU-CTC model showed more robust performance, with low Character Error Rate (CER) and Word Error Rate (WER). This research work aims to support and encourage the research community in developing a robust recognition system for preserving Urdu handwritten literature.
\end{abstract}


\maketitle

\section{Introduction}
The history of the Urdu language with its roots going back to the 12\textsuperscript{th} century, is indeed very rich and fascinating. After the division of British India in 1947, it was declared as Pakistan’s national language \cite{khan2018urdu} and is now widely spoken and written by millions of people in Pakistan, Afghanistan, India, the UAE, and Bangladesh \cite{naz2016offline,rashid2022scrutinization}. Its literary composition began in Deccan during 14\textsuperscript{th} century and was primarily limited to the religious content at that era. After the 15\textsuperscript{th} century, when the use of Urdu spread to the northern regions of India, a large amount of literature emerged, manually written by people \cite{rashid2022scrutinization} locally known as \textbf{Katibs}\footnote{A \textit{\textbf{Katib}} refers to a person who writes in a structured way following calligraphic rules, also known as a \textit{calligrapher}.} \cite{ahmad2016kpti}.

This ancient historic literature is an essential part of the cultural heritage of Urdu-speaking regions. However, the retrieval and preservation of this vast amount of data that has been kept in hard form for centuries is very challenging as well as it remained unexplored to the world \cite{ul2022convolutional-recursive}. Although, a very confined amount of this data is available on internet in the form of images \cite{ganai2022novel-holistic}, but it demands a huge amount of storage space. Additionally, the non-editable nature of text within images makes the information retrieval unfeasible \cite{naz2016offline}. Therefore, making the digital replica of this data will save a lot of valuable resources such as storage space, time, human effort etc. and primarily improves the manageability and accessibility of the material. In this regard, a robust \textbf{Urdu Handwritten Text Recognition} (UHTR) system is an ultimate option that offers a promising solution by transforming this data into digital format (machine-readable/editable form) \cite{rashid2022scrutinization,ganai2022novel-holistic,ul2022convolutional-recursive,riaz2022conv-transformer}.

When it comes to cursive scripts such as Arabic and its derivatives like Urdu, Persian, and Pashto, both \textbf{OCR}\footnote{An \textit{\textbf{Optical Character Recognition} (OCR)} system transforms printed text from images into machine-readable form, while a \textit{\textbf{Handwritten Text Recognition} (HTR)} system does the same for handwritten text.} and \textbf{HTR}\footnotemark[\value{footnote}] systems are less advanced compared to those designed for non-cursive scripts. The research for cursive scripts started at the end of 20\textsuperscript{th} century. The earliest system for the Urdu script dating back to 2003, primarily designed to recognize the individual printed basic characters \cite{pal2003recognition}. Since last few years significant efforts have been made for the development of Urdu OCR systems, achieving accuracies of up to \textbf{98\%} \cite{naz2016offline,khan2018urdu}. In contrast, a very confined amount of work has been done for UHTR \cite{kashif2021urdu-resnet18}, with no commercial UHTR system is available to date \cite{shaiq2022transformer,ganai2023computationallyLRCN,fahadmulti2023mutli-aspect}. This lag is attributed to the unique challenges posed by its script and the lack of standard handwritten datasets, which are discussed in Section \ref{subsec:challenges}.

\subsection{Urdu Script}
\begin{figure}[h!]
  \centering
  \includegraphics[width=12.7cm]{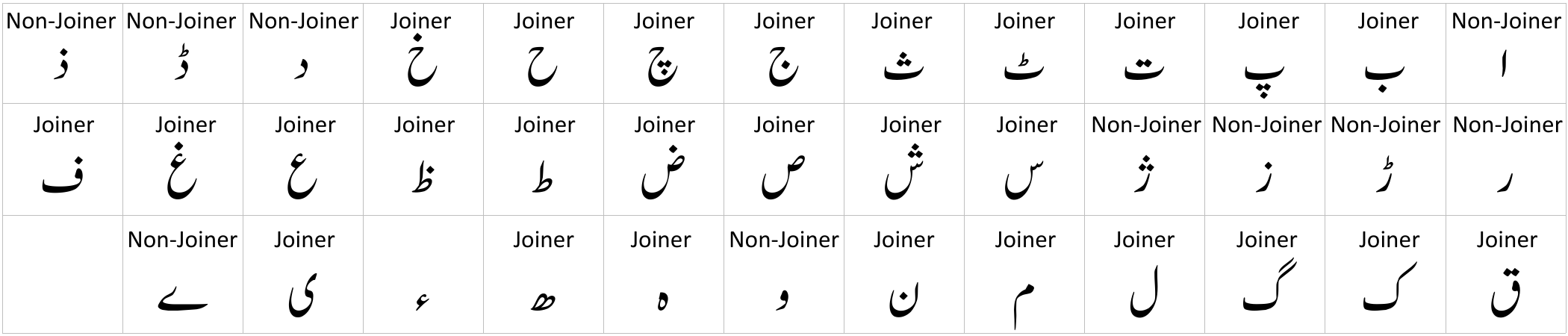}
  \caption{Urdu character set has 38 basic characters including 10 non-joiner characters, 27 joiner characters, and 1 character ‘Hamza’  always occurs isolated \textemdash neither belongs to the joiner class nor to the non-joiner class \protect\cite{khan2018urdu}. \textit{The modern Urdu character set is larger; it has some extensions of the basic characters \protect\cite{satti2012complexities}}.}
  \Description{Representation of Urdu character set along with the tag of class from which they belong, either joiner class or non-joiner class.}
  \label{fig:urdu-characters-set}
\end{figure}

Urdu script has 38 basic characters as shown in Figure \ref{fig:urdu-characters-set}, derived from the Persian alphabets which itself is a super set of Arabic character set; Persian script has 32 characters and Arabic script has 28 characters \cite{sagheer2010holistic,satti2012complexities,khan2018urdu}. Therefore, both Urdu and Persian script adopt the characteristics of Arabic script. Urdu character set consists of two types of characters: joiner characters and non-joiner characters, as shown in  Figure \ref{fig:urdu-characters-set}. A \textbf{non-joiner character} may have only two basic shape forms i.e. \textit{``final''} and \textit{``isolated''}. Alternatively, a \textbf{joiner character} may have four shape forms i.e. \textit{``initial''}, \textit{``middle''}, \textit{``final''}, and \textit{``isolated''} \cite{mukhtar2010experiments,khan2018urdu}. These shape variations of characters are further discussed in Point \ref{item:characters-shapes-variations} of Section \ref{subsec:challenges}. 

\begin{figure}[h!]
  \centering
    \includegraphics[width=10cm]{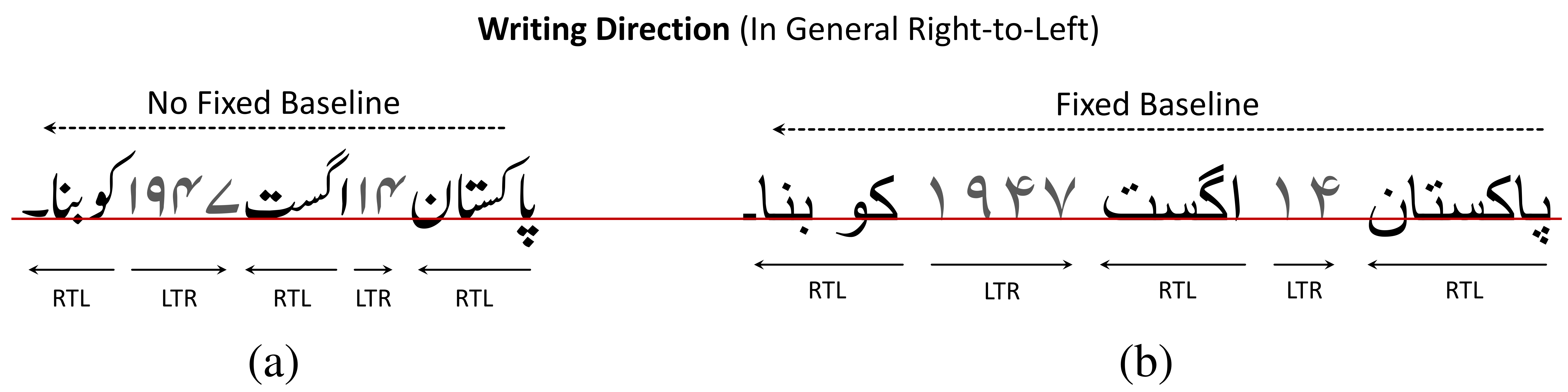}
  \caption{\textbf{(a)} Nastalique Style: Diagonal Behavior (consumes less space), \textbf{(b)} Naskh Style: Horizontal Behavior (consumes more space); often used for writing Urdu but generally used for writing Arabic \protect\cite{hussain2003complexity}. In both (a) and (b), the arrows indicate the writing direction, whereas the red horizontal line shows the \textbf{baseline}\protect\footnotemark.}
  \Description{A sentence ``Pakistan was created on 14th August, 1947.'' is written in Urdu language in two different fonts comprising Nastalique font and Naskh font. The directionality of Urdu script is illustrated using arrows direction, moreover the baselines in both fonts is demonstrated by drawing a horizontal line. Naskh style has fixed baseline whereas Nastalique has no fixed baseline.}
  \label{fig:nastalique-vs-naskh}
\end{figure}

\footnotetext{A \textit{\textbf{baseline}} is an invisible imaginary horizontal line which passes through the text by cutting all the individual characters and ligatures at a certain point \cite{husain2007online}, where the maximum number of pixels are present (this rule is not always true \cite{satti2013offline}).}

Generally, the languages spoken around the worldwide follow a unidirectional writing style while Urdu stands out as a bidirectional language \cite{khan2012urdu}. It means that Urdu characters are written from right-to-left (RTL) whereas numbers are written from left-to-right (LTR) direction \cite{hussain2003complexity,kour2020machine-survey}, as illustrated in Figure \ref{fig:nastalique-vs-naskh}. The standard style for writing Urdu script is \textbf{Nastalique}\footnote{\textbf{\textit{Nastalique}} is a calligraphic style of the `Perso-Arabic' script, developed by an Iranian calligrapher `Mir Ali Heravi Tabrizi' during 14\textsuperscript{th} century \cite{khan2018urdu}. It is a combination of two scripts namely `Naskh' which is used for writing Arabic, and `Taliq' which has been historically used for Persian \cite{hussain2003complexity,rashid2022scrutinization}.} calligraphic style, can be seen in Figure \ref{fig:nastalique-vs-naskh}a.
    
\subsection{Challenges in Urdu Handwritten Text Recognition (UHTR)}
\label{subsec:challenges}
The peculiarities of Urdu script such as its cursive writing style, context sensitivity, and overlapping nature not only make it aesthetically elegant and an artistic essence but also adds significant complexities to its recognition process. Let's see these peculiarities and the associated complexities they entail:

\begin{enumerate}
    \item \textit{Cursive Nature}: Urdu script adopts cursive writing style which means it is written by joining/linking characters together in the form of ligatures, can be seen in Figure \ref{fig:uhtr challenges}a. A word may contain one or more ligatures, and a \textbf{ligature}\footnote{A \textit{\textbf{ligature}} contains one \textit{primary connected component} (longest stroke written without lifting the pen, referred to as primary stroke or main body) and zero or more than zero \textit{secondary connected components} (diacritics)  \cite{lehal2013recognition}.} is composed by combining one or more characters cursively together \cite{hussain2003complexity,satti2012complexities,kour2020machine-survey}.

    \begin{figure}[h!]
      \centering
        \includegraphics[width=10.5cm]{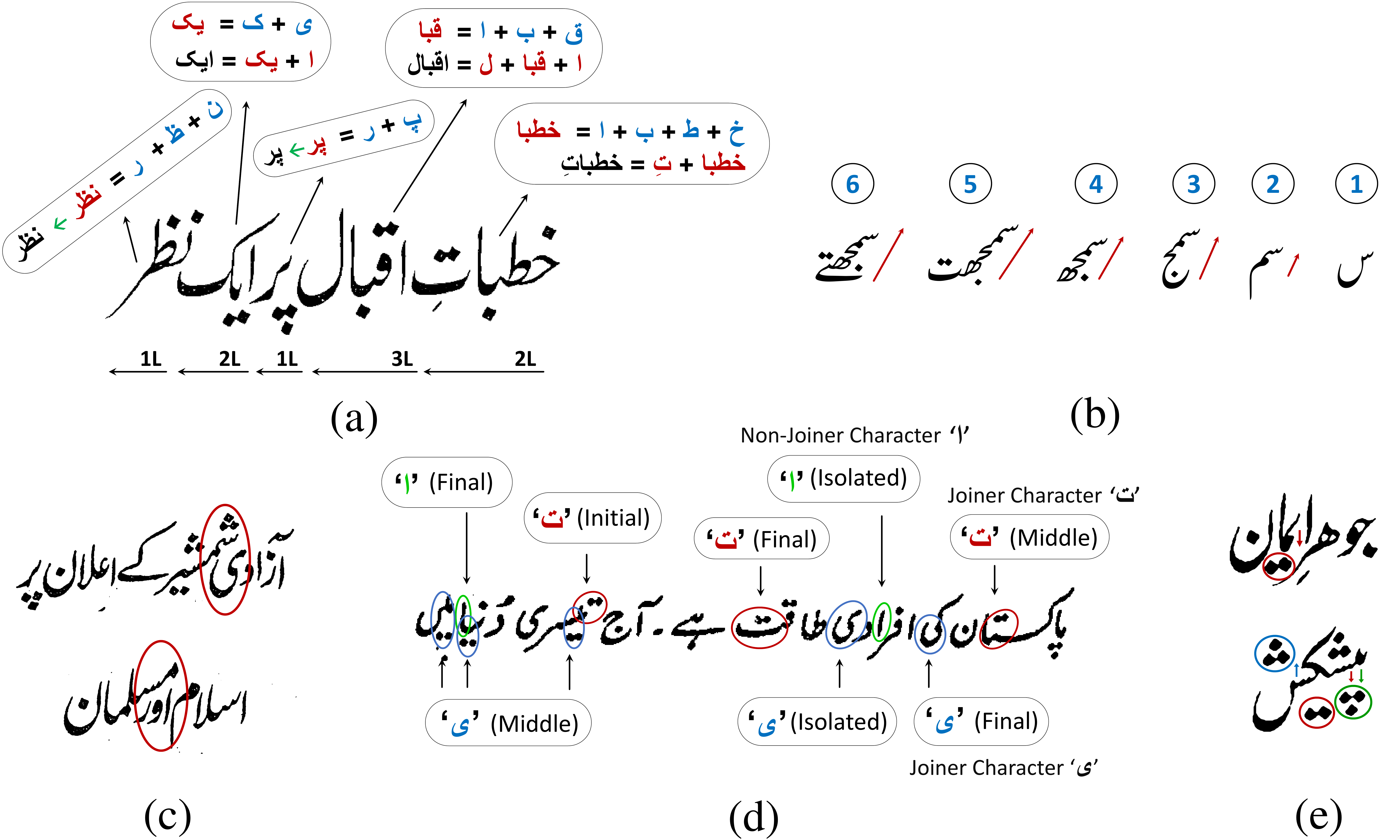}
        \caption{\textbf{(a) Cursive Nature:} The given phrase is made up of five words, each word has one or more ligatures i.e. 1\textsuperscript{st} word has two ligatures (2L) \textemdash first ligature is formed by joining four characters together while second ligature has only one character. In the given example, the arrow above each word points to the details of ligatures present in the corresponding word i.e. individual characters are written in blue color, ligatures in red color, and words in black color. \textbf{(b) Diagonal Behavior:} The process of forming the word is depicted in six steps. In each step a new character is added. As characters are being added, the previous ones are arranged diagonally and progressively shifted towards the upper right corner \protect\cite{muaz2010urdu}. The arrow shows the diagonality and progressive shift of characters towards the top right corner. \textbf{(c) Overlapping Nature:} The oval mark represents overlapping nature; characters vertically overlap with the preceding characters. \textbf{(d) Context Sensitivity:} Context dependent shapes of different characters i.e. non-joiner character: `Alif', and joiner characters: `Tay' and `Choti-Yay'. \textbf{(e) Incorrect Placement of Dot(s):} Arrow points the standard positions of dots, although they are shifted towards left side.}
        \Description{The figure illustrates the cursive nature, diagonal flow, character overlapping, and shape variations of the Urdu script. The images contain text samples written by Katib in the Nastalique calligraphic style.}
        \label{fig:uhtr challenges}
    \end{figure}
    
    \item \textit{Diagonal Behavior}: While writing in Urdu, characters are inclined diagonally from the upper right to the lower left \cite{hussain2003complexity}, that means all the ligatures are tilted at some angle due to this diagonal orientation \cite{javed2009improving}, see Figure \ref{fig:uhtr challenges}b. It helps in consuming less horizontal space as shown in Figure \ref{fig:nastalique-vs-naskh}a, but in turn the baseline does not remain fixed which plays a significant role in detecting skewness.

    \item \textit{Overlapping Nature}: The diagonal and cursive nature of Urdu script causes vertical overlapping between characters i.e. characters can vertically overlap with the preceding characters or ligatures as illustrated in Figure \ref{fig:uhtr challenges}c. It makes the segmentation process complex \cite{husain2007online,satti2013offline,satti2012complexities} \textemdash poses complexities in ligature and character level segmentation.
    
    \item \textit{Context Sensitivity}: Urdu characters change their shapes/visual appearance w.r.t the context in which they occur, commonly referred to as \textbf{context sensitivity} \cite{hussain2003complexity}, see Figure \ref{fig:uhtr challenges}d. A character’s shape is basically dependent upon its connected neighboring characters as well as its position in the ligature, which can be “initial”, “middle”, “final”, or “isolated” \cite{shahzad2009urdu,muaz2010urdu,khan2012urdu,bin2017ucom}. This high context sensitivity of Urdu script complicates accurate character identification and classification.
    \label{item:characters-shapes-variations}

    \begin{figure}[h!]
      \centering
        \includegraphics[width=12cm]{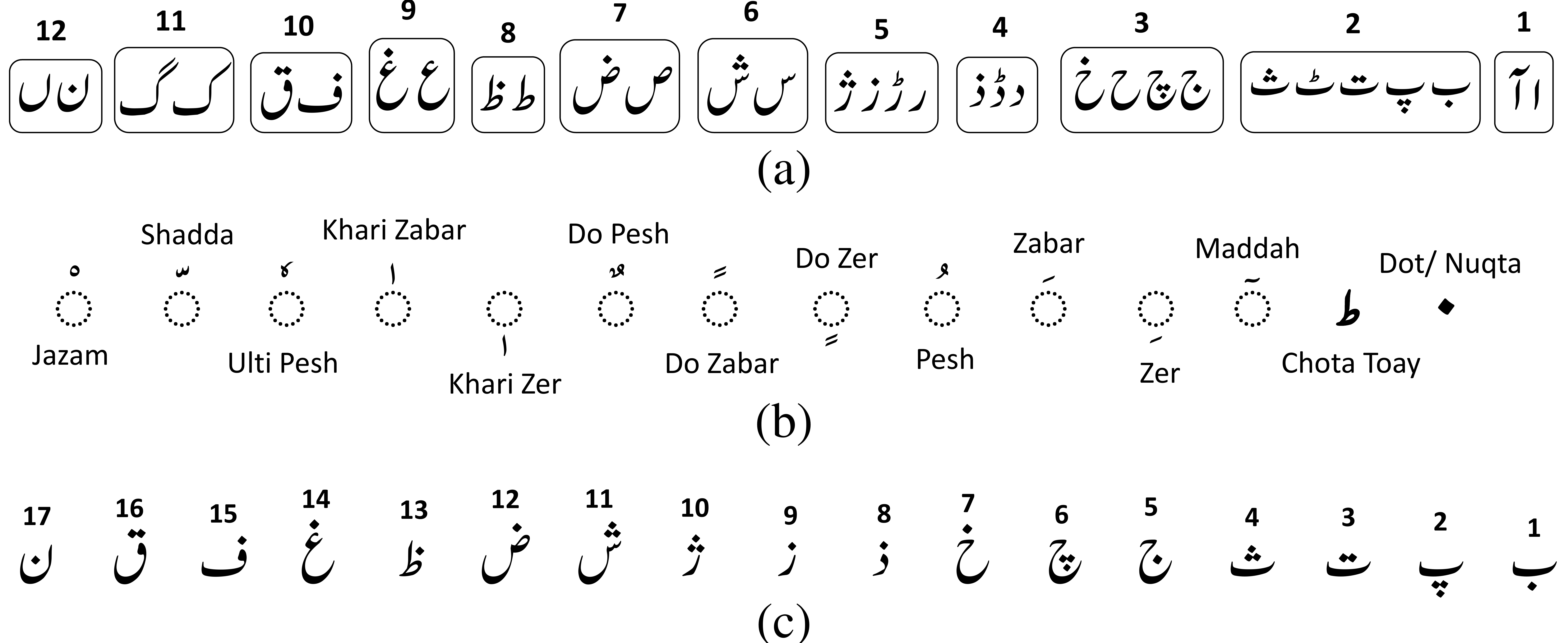}
          \caption{\textbf{(a)} Group of characters having similar primary strokes \protect\cite{kashif2021urdu-resnet18}. \textbf{(b)} There are three primary types of diacritics: “Dot/Nuqta”, “Chota Toay” in superscript, and “Aerabs”. Nuqta and small toay are compulsory diacritics. The remaining diacritics are known as \textit{Aerabs}, which are optional and used for removing any ambiguity in pronunciation \protect\cite{hussain2003complexity,mukhtar2010experiments,satti2012complexities,khan2018urdu}. \textbf{(c)} List of characters accompanied by dots. Dots can range from one to three and can be placed above, center and below of the character’s primary stroke. Different number and placement of dots help in distinguishing characters having similar primary strokes \protect\cite{satti2012complexities,satti2013offline,naz2014optical,khan2018urdu}.}
          \Description{The image illustrates Urdu diacritics, characters sharing similar primary strokes, and characters distinguished by dots}
        \label{fig:diacritics}
    \end{figure}

    \item \textit{Shape Similarities}: Developing a HTR system for a language having structural similarities among its characters is a very complicated task, and Urdu is a prime example of such complexity. Figure \ref{fig:diacritics}a lists the group of characters having similar primary strokes; the thing that differentiates them are the \textbf{diacritical marks}\footnote{A \textit{\textbf{diacritical mark}} is a special type of mark surrounded by the primary stroke, also referred to as \textit{secondary stroke}.} (secondary strokes) given in Figure \ref{fig:diacritics}b. There are total seventeen characters in Urdu character set that are surrounded by dots (compulsory diacritics) \cite{khan2018urdu}, listed in Figure \ref{fig:diacritics}c. 

    \item \textit{Complex/Incorrect Placement of Dot(s)}: The diagonal behavior and overlapping nature of Urdu script makes the placement of dot(s) complex \cite{satti2012complexities}. They might be positioned in such a way where it is difficult to associate each dot (nuqta) to its primary stroke, as shown in Figure \ref{fig:uhtr challenges}e.
\end{enumerate}

In a nutshell, developing an UHTR system is widely regarded as a challenging task, as there is more overlapping, more incorrect/complex placement of diacritics, and more variations in character shapes. Unlike printed text recognition where characters typically have consistent shapes with no or minimal variations. There are considerably more character shape variations in handwritten text, not only due to individuals writing styles but also because the same writer may produce different pen movements when writing the same character. 

Another notable challenge that is associated with UHTR is the scarcity and unavailability of standard datasets \cite{khan2018urdu,kour2020machine-survey,riaz2022conv-transformer,ganai2023computationallyLRCN,ganai2023computationally-efficient,anjum2025survey,alazzawi2026crosslanguagelearningarabicscript}; \textit{\textbf{manually labeling/transcribing a large amount of data is quite challenging}}. There are very few Urdu handwritten datasets that are publicly available for research purposes. Although some of them are not fully accessible, they are partially available which are not enough to train a robust and efficient model for recognition purposes \cite{anjum2020attention}.

Furthermore, several studies \cite{ganai2022novel-holistic,rashid2022scrutinization,ul2022convolutional-recursive,riaz2022conv-transformer}, highlighted the potential of Urdu handwriting recognition which could contribute to the preservation of historical literature for eternity. However, until now they did not specifically focus on this particular type of writing that is \textit{flat nib writing}. \textbf{They did work on the recognition of simple pen/ballpoint writing whereas the historical literature has been mostly written in flat nib writing}.

\subsection{Our Contributions}
\label{subsec:contributions}
To advance research in the domain of UHTR, our main contributions of this paper are as follows:

\begin{enumerate}
    \item[$\bullet$] In this work, a specialized offline text lines dataset called the \textbf{Urdu Katib Handwritten Dataset} (UKHD) is presented for the research community.
    \item[$\bullet$] The introduced dataset comprises flat nib writing in nastalique calligraphic style, \textit{written by experts in historical times}. It will support the research community to develop a robust recognition system to preserve the Urdu handwritten literature.
    \item[$\bullet$] Semi-automatic approaches for segmenting and labeling the text line images are introduced, which leverage existing methods and techniques. These approaches significantly reduce the time and effort required for dataset creation.
    \item[$\bullet$] Additionally, the effectiveness of different CRNN-based hybrid models has been evaluated on the primary subset of UKHD to report the baseline results and optimal architecture for \textbf{Urdu Katib Handwriting Recognition} (UKHR).    
\end{enumerate}

The rest of the paper is structured as follows: Section \ref{sec:LR} gives a detailed review of existing work that has been done for UHTR. Section \ref{sec:ukhd} provides UKHD statistics, subsequently section \ref{sec:ukhd generation} describes a comprehensive account of methods and techniques employed for its generation. Section \ref{sec:implementation} presents the implementation of hybrid models, covering a detailed explanation of model architecture, experimental data, data preparation, the meticulous training process and performance evaluation. Then, section \ref{sec:results} unveils the outcomes of the evaluated hybrid models and provides insightful discussion of the optimal model output and analysis of its best and failure cases. Ultimately, section \ref{sec:conclusion} concludes the study and suggests the directions for future work.

\section{Literature Review}
\label{sec:LR}
In the realm of Urdu script recognition, two types of recognition techniques are being followed in literature: the \textbf{holistic} approach and the \textbf{analytical} approach. In holistic approach, text recognition occurs at word or sub-word/ligature level. It is also referred to as \textbf{segmentation-free} approach as ligatures are not further segmented into characters. Alternatively in analytical approach, the text recognition takes place at character level, which is also known as \textbf{segmentation-based} approach \cite{javed2010segmentation,satti2013offline}. It is further divided into two strategies: explicit segmentation-based recognition and implicit segmentation-based recognition. In \textbf{explicit} segmentation-based recognition, the ligature is further segmented into characters or smaller units/primitives explicitly using some heuristics or predefined rules. While in \textbf{implicit} segmentation-based recognition, predefined labels (transcriptions) correspond to text images, and the model automatically learns the segmentation points during the recognition process \cite{naz2014optical,naz2016offline,khan2018urdu}.

Mukhtar et al. \cite{mukhtar2010experiments} adopted a holistic approach and introduced the very first Urdu handwritten word recognition system using a dataset of 1,600 instances of 100 words from two writers. They extracted Gradient Structural Concavity (GSC) features from normalized word images, resulting in 512-bit binary vectors that were then classified using kNN and SVM classifiers, yielding accuracies of 70\% and 75\% respectively. Sagheer et al. \cite{sagheer2010holistic} used compound feature sets comprising gradient and structural features, with SVM classifier, and achieved a notable recognition accuracy of 97\% on the CENPARMI Urdu Words Database. In 2021, Shah et al. \cite{shah2021urdu} presented a CNN-based model, leveraging transfer learning with the MobileNet architecture. The proposed system was evaluated on their custom generated dataset (603 samples of Urdu handwritten and printed words) and obtained a recognition accuracy of 90\%.

In 2022, Ganai \& Khursheed \cite{ganai2022novel-holistic} presented an unconstrained holistic approach for UHTR, utilizing ligatures as a fundamental unit. They introduced the Urdu Handwritten Ligature Dataset (UHLD) comprising 6,000 text lines, having 2,100 unique ligatures. Ligatures were extracted using projection profile-based methods from both datasets \textemdash they also considered Urdu Nastalique Handwritten Dataset (UNHD) \cite{ahmed2019handwritten-blstm} for unique ligature extraction. The separation of primary and secondary components of ligatures was done by their own proposed algorithm, which were then classified using CNNs. Among the evaluated CNN variants VGG-Net outperforms, gave 93\% recognition on UNHD and 97\% on UHLD. In 2023, they introduced another holistic approach for Urdu handwriting recognition using LRCN model. This involved the creation of 1,500 ligature classes derived from two benchmark datasets: 500 from UNHD \cite{ahmed2019handwritten-blstm}, 1000 from UHLD \cite{ganai2022novel-holistic}, and concurrently the recognition/classification of these classes. CNN was utilized for feature extraction, while MDLSTM was employed for classification, achieved an impressive recognition rate of 94.2\% for UNHD and 96.6\% for UHLD \cite{ganai2023computationallyLRCN}.

In holistic approaches, each ligature corresponds to a distinct class, resulting in a high number of classes, far exceeding the total number of Urdu characters and their various forms \cite{naz2016offline}. In Urdu, there are approximately 26,000 unique ligatures, accommodating this large class count is quite challenging. Researchers have attempted to address this by focusing only on the most frequently used ligatures \cite{hassan2019cursive-blstm-casestudy}. However, segmentation-based approaches emerge as a more favorable alternative as they effectively accommodate the huge number of ligature classes by recognizing text at character level.

A Language Independent Optical Character Reader (LIOCR) based on explicit segmentation-based approach, developed by Ali et al. \cite{ali2004language} used thirteen basic geometric strokes of characters for HTR. After performing different preprocessing steps (binarization, baseline removal, thinning), the ligatures were isolated by traversing the image. A segmentation map was used to determine the primitive-level segmentation points, and post-segmentation was further employed to recombine the incorrectly segmented parts. For recognition, a set of stroke features were computed and fed into a trained neural network. An XML file was used as a classifier to categorize each recognized stroke into a character. The system, evaluated on English, Pitman Shorthand Language (PSL), and Urdu, performed relatively poorly with Urdu, with a recognition rate of 70-80\% and several other limitations.

Explicit segmentation-based recognition approaches face challenges in directly segmenting ligatures into smaller units such as characters or primitives, and in identifying the hierarchical composition of these primitives \cite{satti2013offline}, requiring a comprehensive knowledge of the starting and ending points of characters \cite{khan2018urdu}. To mitigate these challenges, implicit segmentation-based recognition techniques are employed, where the model automatically handles segmentation during the recognition process. This approach has shown promising outcomes in achieving better recognition rates for various scripts \cite{shi2016end-to-end,tong2020ma,gader2022attention-arabic} including Urdu printed text recognition \cite{naz2016urdu-implicit,ul2013offline,hassan2019cursive-blstm-casestudy}, as it exploits deep learning algorithms i.e. CNNs, RNNs. Therefore, this approach is being adopted for UHTR. The availability of a large data set becomes imperative for its effective implementation.

Bin Ahmed et al. \cite{bin2017ucom} generated the first Urdu handwritten text lines dataset, the ‘UCOM dataset’, which contains 6,400 text lines penned by 100 Urdu native writers. They traversed a fixed-sized window of 30x1 over the text line images to capture pixel values as feature values for the RNN classifier. The authors reported an error rate of 4$\sim$6\% on a subset of the dataset (50 training and 20 testing text lines). Later, they further extended their work by presenting the `Urdu Nastalique Handwritten Dataset' (UNHD) with 10,000 text lines written by 500 writers. The dataset was partitioned into training, validation, and test sets in 50\%, 30\%, and 20\% ratios, respectively, and evaluated using a BLSTM model, achieving a Character Error Rate (CER) of approximately 6.04 to 7.93\% \cite{ahmed2019handwritten-blstm}. Unlike previous methods that utilized raw pixel value features \cite{bin2017ucom,ahmed2019handwritten-blstm}, Hassan et al. \cite{hassan2019cursive-blstm-casestudy} employed CNN for feature extraction and BLSTM followed by CTC for classification and transcription generation. This model achieved an average Character Recognition Rate (CRR) of 83.69\% when evaluated on 1,000 text lines from their custom dataset, which includes 6,000 lines written by 600 writers.

Anjum and Khan \cite{anjum2020attention} proposed an attention-based encoder-decoder framework for UHTR, employing DenseNet in the encoder for high-level feature extraction and Gated Recurrent Unit (GRU) in the decoder to convert these features into a sequence of characters. The attention mechanism in the decoder focuses on relevant image regions to generate individual characters. They achieved an accuracy of 77.05\% at character level and 43.35\% at word level, when evaluated on their custom generated the PUCIT-Offline Urdu handwritten text lines dataset, which comprises 7,309 text lines with 78,870 words written by 100 writers. Shaiq et al. \cite{shaiq2022transformer} explored a transformer-based model, using the PULT-Offline dataset \cite{anjum2020attention} in their experiments. After image preprocessing and feature extraction with ResNet-18, the extracted features were input into the transformer model. However, the achieved CER was more than 85\% primarily due to the limited dataset.

In \cite{ul2022convolutional-recursive}, a CRNN hybrid model (CNN-BLSTM-CTC model) with tailored enhancements was introduced. The four variants of the proposed model were evaluated by changing the CNN layers, reported a CER of 7.35\% which improved to 5.49\% with the incorporation of an n-gram language model. The experiments were conducted on their presented the `NUST-UHWR' dataset, an unconstrained Urdu handwritten text lines dataset collected from seven domains by the contribution of 1,000 individuals. Riaz et al. \cite{riaz2022conv-transformer} proposed a convolutional transformer-based model that effectively removed the necessity for a separate language model as used in \cite{ul2022convolutional-recursive}. They trained the proposed model on a mix of Urdu handwritten (`NUST-UHWR') and printed (`Urdu Ticker Dataset' and `UPTI-2') datasets, and achieved a CER of 5.31\% when evaluated on 1,061 text lines from UHWR dataset.

So far, the above literature review has shown us the progress made in UHTR, and the challenges of both recognition approaches. However, the adoption of implicit segmentation-based recognition strategies has shown promising outcomes, as they use the power of \textit{hybrid models} that combine the strengths of multiple deep learning models into an end-to-end system. Therefore, this research evaluates various CRNN-based hybrid models on the proposed dataset.

\section{Urdu Katib Handwritten Dataset (UKHD)}
\label{sec:ukhd}
This research presents the \textit{Urdu Katib Handwritten Dataset (UKHD)}, an offline Urdu handwritten dataset containing flat nib writing variations in nastalique calligraphic style. It consists of text line images and their corresponding transcriptions, organized into two primary subsets: the `\textbf{Plain Urdu Text Lines}' (PUTL), and the `\textbf{Mixed Urdu Text Lines}' (MUTL). The PUTL subset exclusively features the Urdu language whereas the MUTL subset has a mix of languages incorporating Urdu, English, Arabic, and Persian. The proposed dataset has the highest number of text lines among existing offline Urdu handwritten text line datasets, can be seen in Table \ref{tbl:existing-urdu-datasets}.

\begin{table*}[h!]
  \caption{Existing Offline Urdu Handwritten Text Line Datasets}
  
  \label{tbl:existing-urdu-datasets}
  \centering
  \begin{tabular}{p{6.2cm}>{\centering\arraybackslash}p{2.8cm}>{\centering\arraybackslash}p{2.8cm}}
    \toprule
    Dataset & Flat Nib Writing  & No. of Text Lines\\
    \midrule
    UCOM dataset \cite{bin2017ucom} & \textcolor{Maroon}{\ding{55}} & 6,400 \\

    UNHD dataset (UCOM extended dataset) \cite{ahmed2019handwritten-blstm} &  \textcolor{Maroon}{\ding{55}} & 10,000 \\
    
    Custom dataset \cite{hassan2019cursive-blstm-casestudy} & \textcolor{Maroon}{\ding{55}} & 6,000 \\

    PUCIT dataset \cite{anjum2020attention} & \textcolor{Maroon}{\ding{55}} & 7,309   \\
    
    ULHD dataset \cite{ganai2022novel-holistic} & \textcolor{Maroon}{\ding{55}} & 6,000 \\
    
    NUST-UHWR dataset \cite{ul2022convolutional-recursive}  & \textcolor{Maroon}{\ding{55}} & 10,608 \\

    \rowcolor{lightblue}
    \hyperref[tbl:ukhd-stats]{\textbf{\textcolor{black}{Proposed dataset (UKHD)}}} & \textcolor{teal}{\ding{51}} & \textbf{13,213} \\
    \bottomrule
  \end{tabular}
\end{table*}

\begin{table}[h!]
  \caption{Details of the Iqbaliyat (Iqbal Studies) Books Used for UKHD Dataset Creation (\textit{Book titles are hyperlinked to their corresponding PDFs.})}
  \label{tbl:books-info}
  \begin{tabularx}{\textwidth}{>{\centering\arraybackslash}p{1.5cm}>{\centering\arraybackslash}p{0.8cm}>{\centering\arraybackslash}p{5.7cm}>{\centering\arraybackslash}p{1cm}>{\centering\arraybackslash}X}
    \toprule
    Book ID & Year & Book Title & Pages & Content Layout \\
    \midrule
    
    001 & 1883 & \href{https://iqbalcyberlibrary.net/en/Khutbat-e-Iqbal-par-ek-nazar.html}{Khutbat-e-Iqbal Par Aik Nazar} & 88 & Nasar\\
    
    002 & 1977 & \href{https://iqbalcyberlibrary.net/en/Iqbal-aur-Tisree-Dunya-Kausar-Niazi.html}{Iqbal Aur Teesri Duniya} & 53 & Nasar\\
    
    003 &  1939 & \href{https://iqbalcyberlibrary.net/en/Allama-Iqbal-Muhammad-Hussain.html}{Allama Iqbal} & 111 & Nasar + Nazm \\
    
    004 & 1961 & 
    \href{https://iqbalcyberlibrary.net/en/Falsafa-e-Islam-aur-Iqbal-Al-islam-aur-Iqbal-Meer-Muhammad-Khan.html}{Fasoos Al-Islam Aur Iqbal} & 133 & Nasar \\
    
    005 & 1966 & \href{https://iqbalcyberlibrary.net/en/Baqiyat-e-Iqbal-Moeeni-1966.html}{Baqiyat-e-Iqbal} & 497 & Nazm + Ghazal \\
    
    006 & 1936 & \href{https://www.iqbalcyberlibrary.net/en/1922.html}{Zarb-e-Kaleem} & 190 & Nazm + Ghazal \\
    
    \bottomrule
  \end{tabularx}
\end{table}

\begin{table}[h!]
  \caption{UKHD Statistics Regarding Extracted Text Lines from Each Source Book}
  \label{tbl:ukhd-stats}
  \centering
  \begin{tabularx}{\columnwidth}{>{\centering\arraybackslash}p{1.3cm}>{\centering\arraybackslash}p{3.2cm}>{\centering\arraybackslash}p{3cm}>{\centering\arraybackslash}X}
    \toprule
    Book ID & Extracted PUTL & Extracted MUTL & Total Extracted Text Lines \\
    \midrule
    001 & 1,642 & 165 & 1,807 \\
    002 & 689 & 51 & 740 \\
    003 & 1,540 & 69 & 1,609 \\
    004 & 1,662 & 345 & 2,007 \\
    005 & 4,446 & 806 & 5,252 \\
    006 & 1,763 & 35 & 1,798 \\
    \midrule
    \rowcolor{lightblue}
    & Total \hyperref[fig:putl-hist]{\textcolor{black}{PUTL = 11,742}} & Total 
    \hyperref[fig:mutl-hist]{\textcolor{black}{MUTL = 1,471}} & Total text lines in UKHD = 13,213 \\
    \bottomrule
  \end{tabularx}
\end{table}

\begin{figure}[h!]
    \centering    \includegraphics[width=12.7cm]{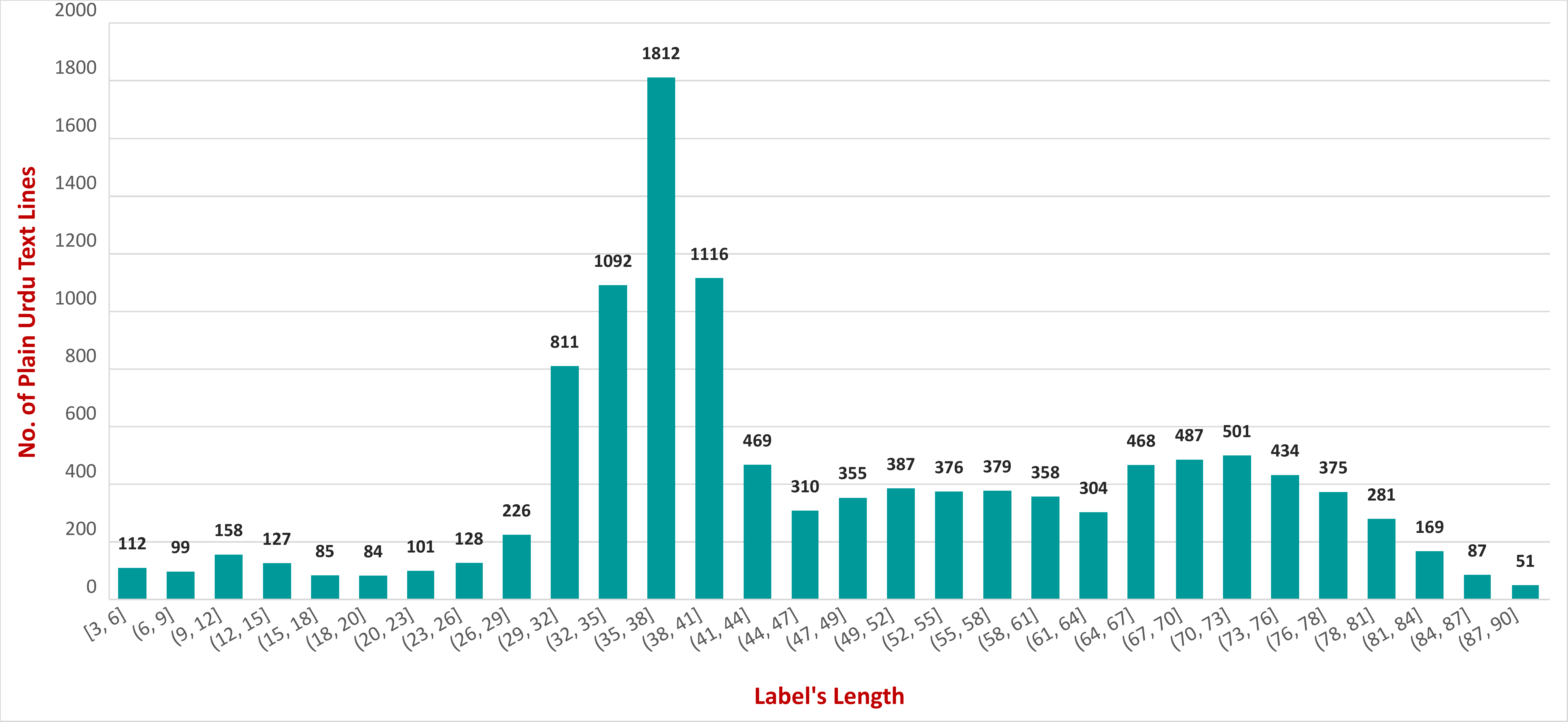}
    \caption{PUTL Subset Labels Length Histogram \textemdash  Minimum Length is 3, whereas Maximum Length is 90}
    \Description{histogram that shows the labels length of plain Urdu text line images subset}
    \label{fig:putl-hist}
\end{figure}

\begin{figure}[h!]
  \centering     
  \includegraphics[width=12.7cm]{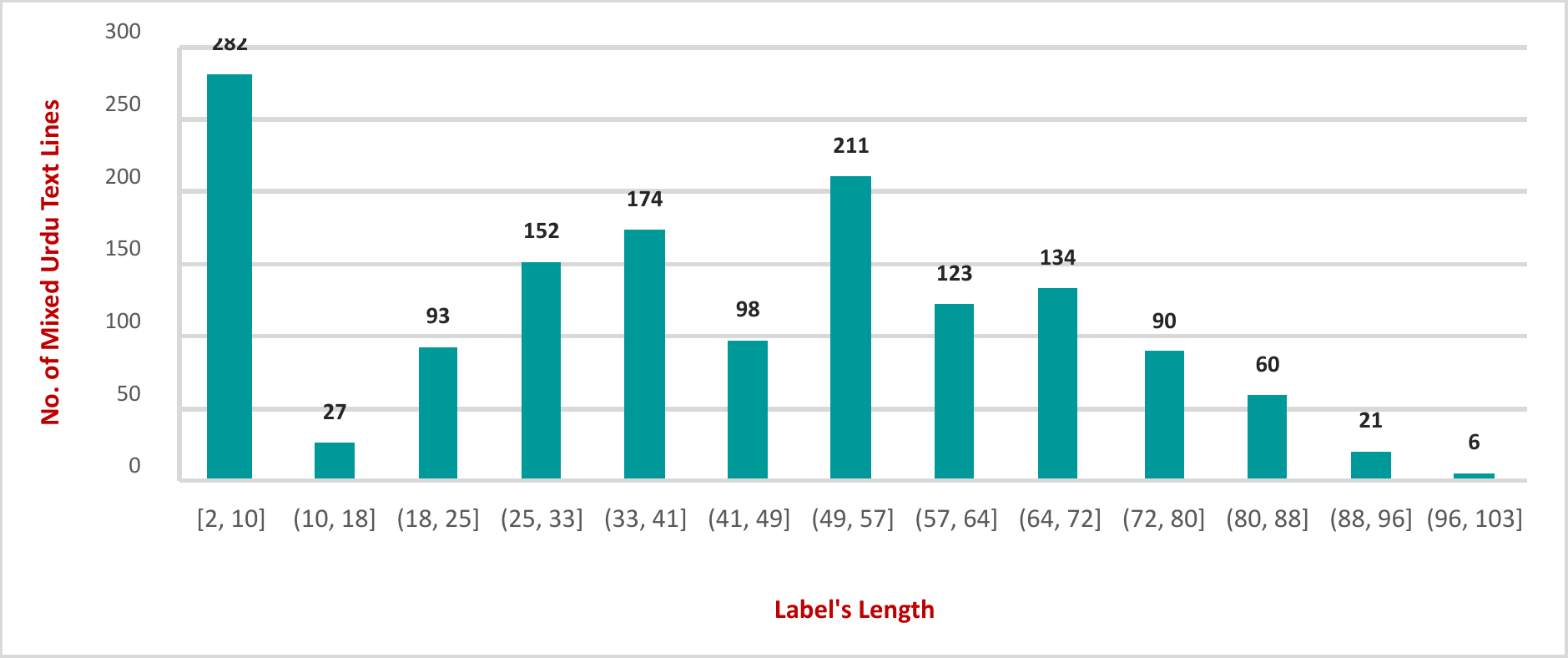}
  \caption{MUTL Subset Labels Length Histogram \textemdash Minimum Length is 2, whereas Maximum Length is 103}
  \Description{histogram that shows the labels length of mixed Urdu text line images subset}
  \label{fig:mutl-hist}
\end{figure}

The detailed information about the books used for UKHD creation is provided in Table \ref{tbl:books-info}. The statistics of UKHD regarding extracted text lines from each source book are presented in Table \ref{tbl:ukhd-stats}. Additionally, the histograms illustrating the \textit{labels length}\footnote{The \textit{\textbf{label length}} refers to the number of characters in a text line (label).} for both subsets of UKHD, the PUTL and MUTL are depicted in Figure \ref{fig:putl-hist} and \ref{fig:mutl-hist} respectively.

\section{Methodology for UKHD Generation}
\label{sec:ukhd generation}
\begin{figure}[h!]
  \centering     
    \includegraphics[width=11.2cm]{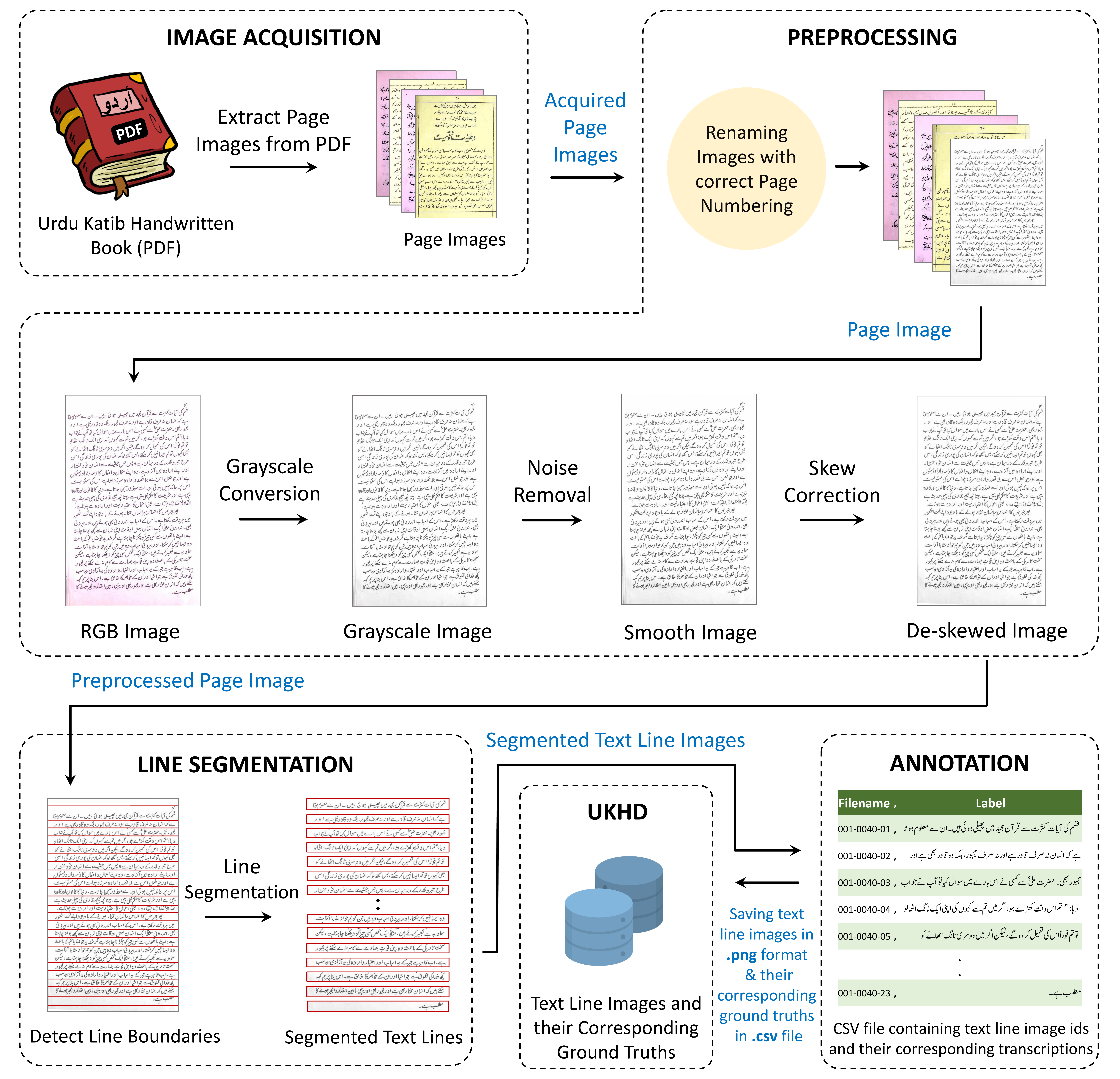}
  \caption{UKHD Generation Process}
  \Description{A diagram that represents UKHD generation process}
  \label{fig:ukhdg}
\end{figure}

The overall UKHD generation process is elaborated in Figure \ref{fig:ukhdg} comprising four distinct phases i.e. image acquisition, preprocessing, line segmentation and annotation that collectively contribute to its generation.

\subsection{Image Acquisition}
\label{sec:image-acquisition}
\begin{figure}[h!]
  \centering      
  \includegraphics[width=9.6cm]{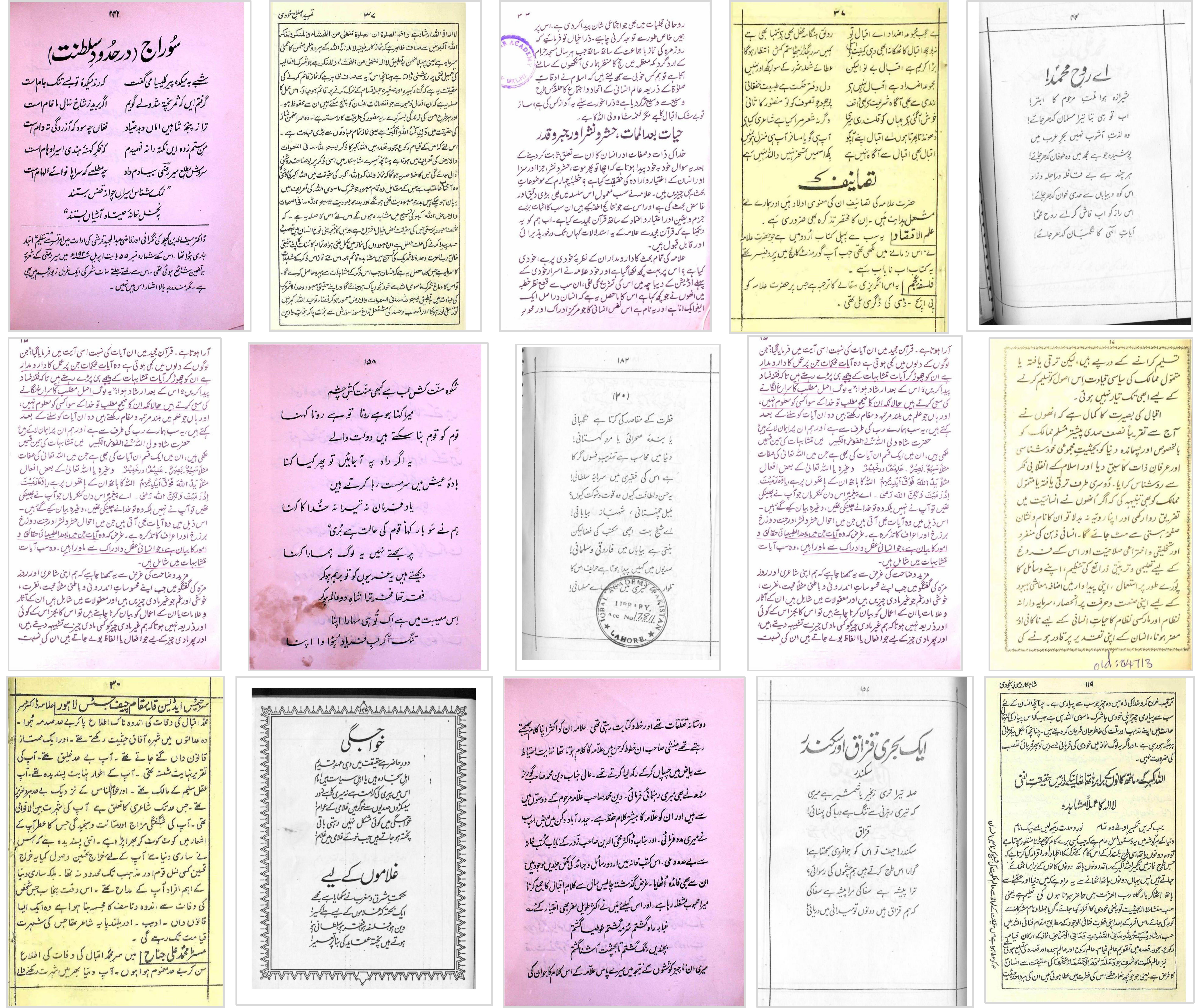}
  \caption{Sample Images from Source Books used in UKHD/ Urdu Katib Handwriting Samples}
  \Description{}
  \label{fig:source-books-samples}
\end{figure}

The text images were digitally acquired from Urdu calligrapher/katib written materials. The data source consists of six Urdu books written by katibs in old times, having flat nib writing in nastalique calligraphic style. The books were downloaded in PDF format from the \href{https://iqbalcyberlibrary.net/}{\textit{Iqbal Cyber Library}} \cite{iqbalcyberlibrary}, the official digital repository of \textbf{Iqbal Academy Pakistan}. The page images were subsequently extracted using an online PDF-to-image conversion tool. Detailed information about the books is provided in Table \ref{tbl:books-info}, and sample images are shown in Figure \ref{fig:source-books-samples}.

\subsection{Preprocessing}
Page images were first renamed using correct page numbers for consistency, then the following preprocessing steps were applied to improve image quality and reduce computational complexity.

\subsubsection{Grayscale Conversion}
\begin{figure}[h!]
  \centering     
  \includegraphics[width=7cm]{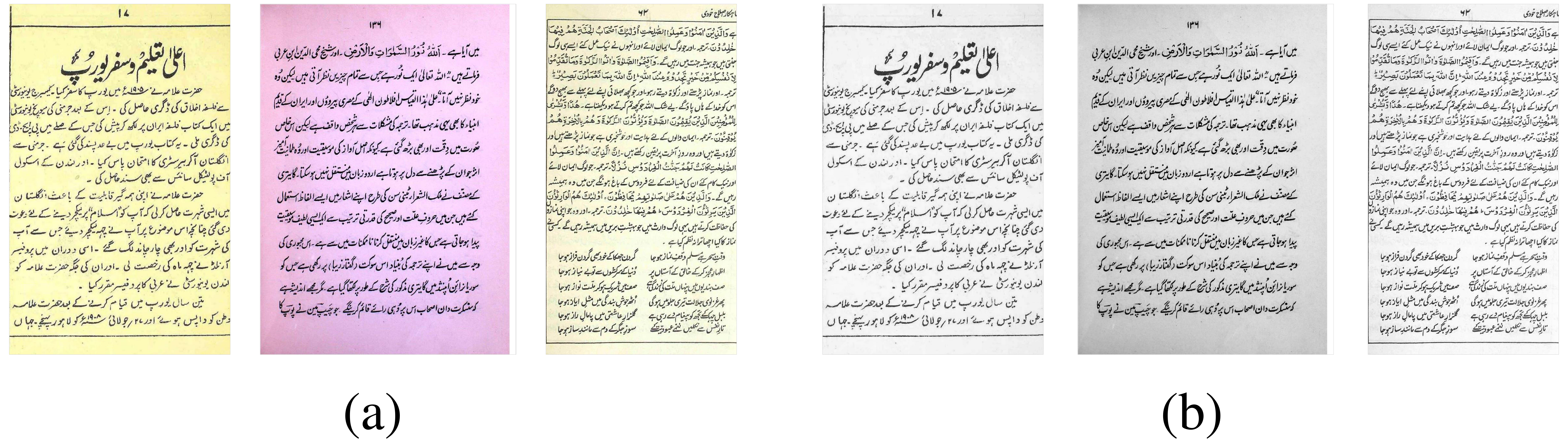}
  \caption{\textbf{(a)} Samples of Acquired RGB Images \textbf{(b)} After Grayscale Conversion}
  \Description{Samples of RGB images and their grayscale conversions. RGB images are colorful whereas grayscale images are black and white.}
  \label{fig:rgb-vs-gray}
\end{figure}

During text recognition, the structure of the text is important. Therefore, the color information of the acquired RGB images was eliminated by converting them into grayscale color space using the python library ‘\textbf{OpenCV}’. Samples are shown in Figure \ref{fig:rgb-vs-gray}, the resultant grayscale images are computationally less complex as well as they retained sufficient information about the structure of text.

\subsubsection{Noise Removal}
The books used for UKHD creation are quite antiquated; consequently, some images had text shadows from the reverse side of the pages. To make the images smooth while preserving the structural details of the text, the \textbf{Median}\footnote{In \textbf{\textit{median filtering}}, each pixel’s value is replaced with the median value of its neighboring pixels within a window/kernel.} filter has been applied. As the level of noise varied across the books, therefore adjustments to the filter \textit{window size}\footnote{The \textit{\textbf{window size}} is used to determine the number of neighboring pixels to consider while calculating the median of each pixel’s value. Smaller size removes minor disturbances while large size eliminates larger noise patterns.} were made according to the condition of each book as it directly influences the balance between eliminating and preserving the details in the image \textemdash chosen window sizes were 3, 5, and 7.

\begin{figure}[h!]
  \centering     
  \includegraphics[width=11cm]{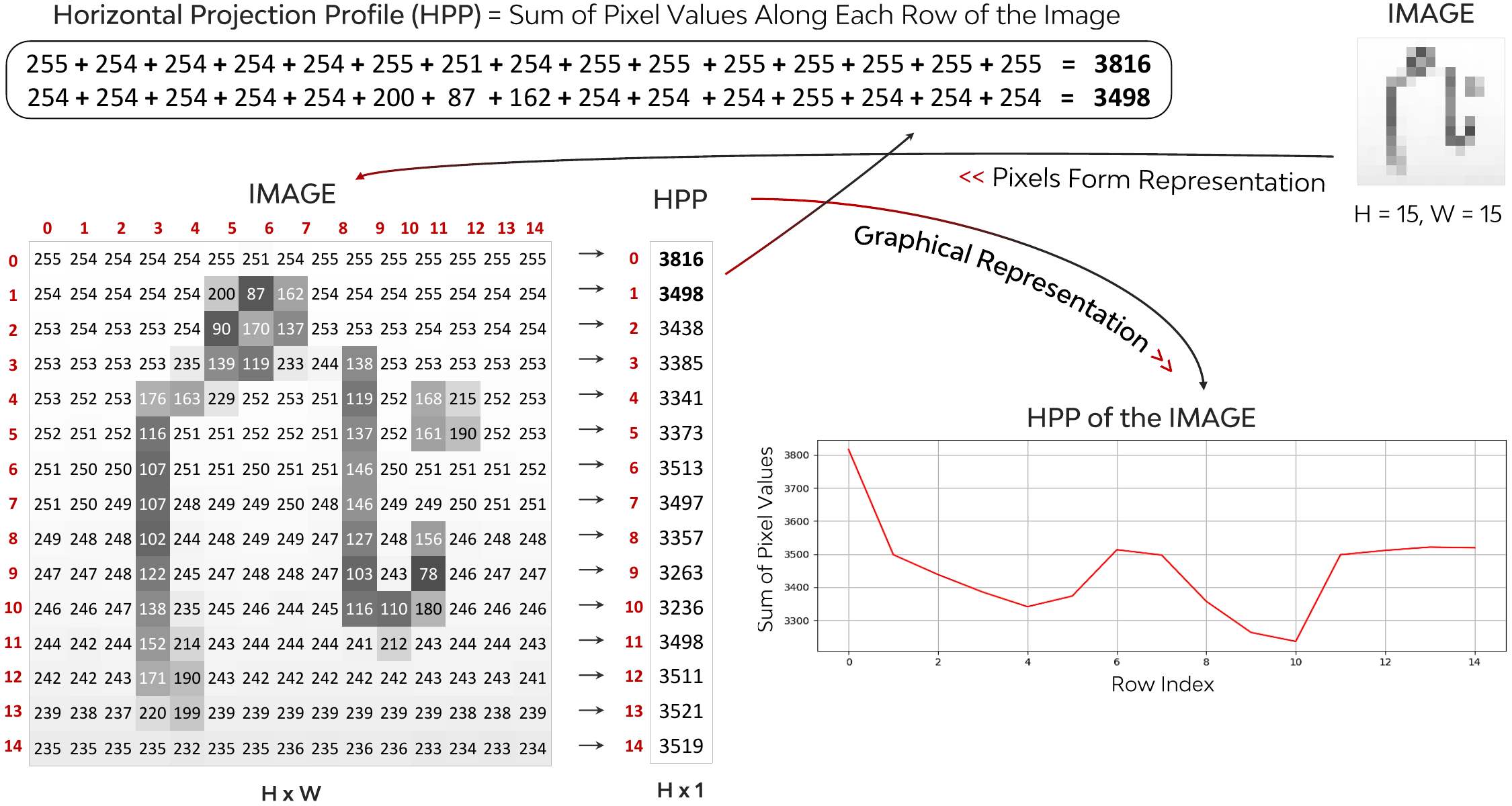}
  \caption{Horizontal Projection Profile (HPP) \textemdash Column Vector Hx1 is the HPP of the Image HxW (It converts a 2D image into 1D signal)}
  \label{fig:hpp}
  \Description{In this figure, horizontal projection profile of an image is represented. Image in pixels form and HPP in graphical form is also illustrated.}
\end{figure}

\subsubsection{Skew Correction}
Following that, the orientation of the image has been corrected to ensure accurate line segmentation. There were many images in which text lines were not perfectly aligned to the baseline, they were distorted at an angle either positively skewed or negatively skewed. A \textbf{Horizontal Projection Profile} (HPP)\footnote{The \textit{\textbf{Horizontal Projection Profile} (HPP)} is an accumulated sum of pixel values along the x-axis of an image, also known as \textit{horizontal histogram} \cite{javed2009improving,mahanta2013skew}, as illustrated in Figure \ref{fig:hpp}. It is a technique used to examine the distribution of pixel intensities along the horizontal axis of an image.} based method has been employed to make them zero-skewed, its pseudo code is given in Algorithm \ref{algo:skew-correction}. It automatically determines and applies the optimal rotation angle for deskewing the given input image.

\begin{algorithm}[h!]    
    \small
    \SetAlgoNlRelativeSize{0}
    \SetKwInput{KwInput}{Input}
    \SetKwInput{KwOutput}{Output}
    
    \KwInput{skewed\_img}
    \KwOutput{deskewed\_img}
    
    \SetKwFunction{FMain}{skew\_correction}
    \SetKwProg{Fn}{Function}{:}{}
    
    \BlankLine
    \Fn{\FMain{skewed\_img}}{
        \footnotesize
        $trans\_skewed\_img \gets invert(sobel(skewed\_img))$ 
        
        $predicted\_angle \gets 0$
        
        $highest\_median \gets 0$ 
        \BlankLine
        \small
        \For{angle in range (-5, 5)}{
            \footnotesize $rotated\_img \gets rotate(trans\_skewed\_img,  angle)$\;
            $hpp \gets sum(rotated\_img,  x\_axis)$\;
            $hpp\_median \gets median(hpp)$\; 
            \BlankLine
            \small
            \If{highest\_median $<$ hpp\_median}{
                \footnotesize
                $predicted\_angle \gets angle$\;
                $highest\_median \gets hpp\_median$\;
            }
        }
        \BlankLine
        \footnotesize
        $deskewed\_img \gets rotate(skewed\_img, predicted\_angle)$\;

        \small
        \KwRet{deskewed\_img}\;
    }
\caption{Skew Correction Algorithm}
\label{algo:skew-correction}
\end{algorithm}

\begin{figure}[h!]
      \centering            
      \includegraphics[width=11cm]{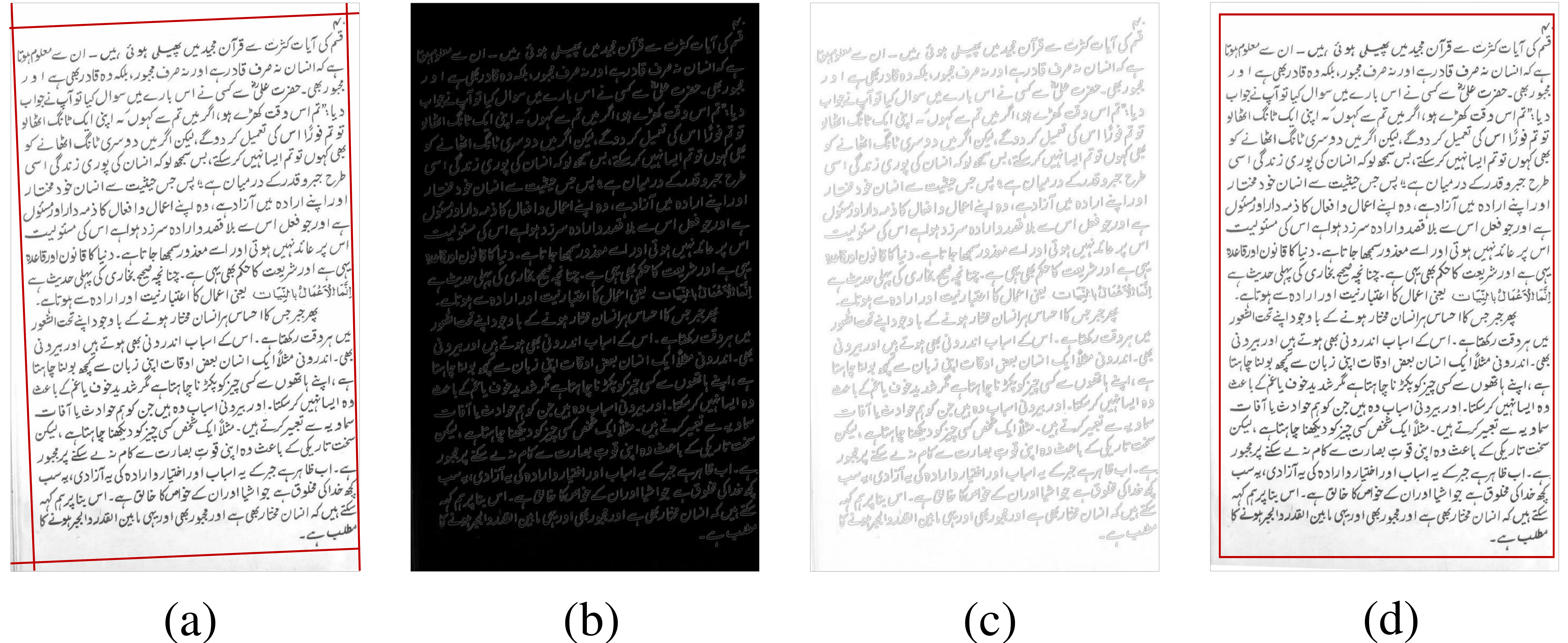}
      \caption{\textbf{(a)} Noise Free Grayscale Skewed Image, \textbf{(b)} After Applying Sobel Filter: It detected horizontal and vertical edges, and eliminated the extra information such as the gradual changes in intensities which are not associated with significant edges. \textbf{(c)} After Performing Image Inversion/Negation: It enhanced the visibility of the text. It is transformed skewed image which has been subsequently rotated at different angles. \textbf{(d)} De-skewed Image: After rotating the skewed image at determined optimal rotation angle i.e. ‘-2’.}
      \Description{}
      \label{fig:skewness-to-deskewness}
    \end{figure}

The step-by-step working of this algorithm using an example along with visual representations is as follow:

\begin{enumerate}
    \item Initially, the grayscale skewed image given in Figure \ref{fig:skewness-to-deskewness}a is processed to enhance the text features within the image. The `Sobel' filter first detects the boundaries/edges of the text and eliminates irrelevant intensity changes, as shown in Figure \ref{fig:skewness-to-deskewness}b. Then inversion operation further highlights the text features, as depicted in Figure \ref{fig:skewness-to-deskewness}c, the resulting image is termed as \textit{transformed skewed image}.  

    \item Following preprocessing, the transformed skewed image given in Figure \ref{fig:skewness-to-deskewness}c is iteratively rotated at ten different angles, ranging from [-5, 5). The resultant images are termed as the \textit{rotated images}. The Horizontal Projection Profile (HPP) of each rotated image is then calculated that provides information about the distribution of pixel intensities along their horizontal axis, their visual representations are illustrated in Figure \ref{fig:rotated-imgs-hpps}.

    \begin{figure}[h!]
      \centering     
      \includegraphics[width=12cm]{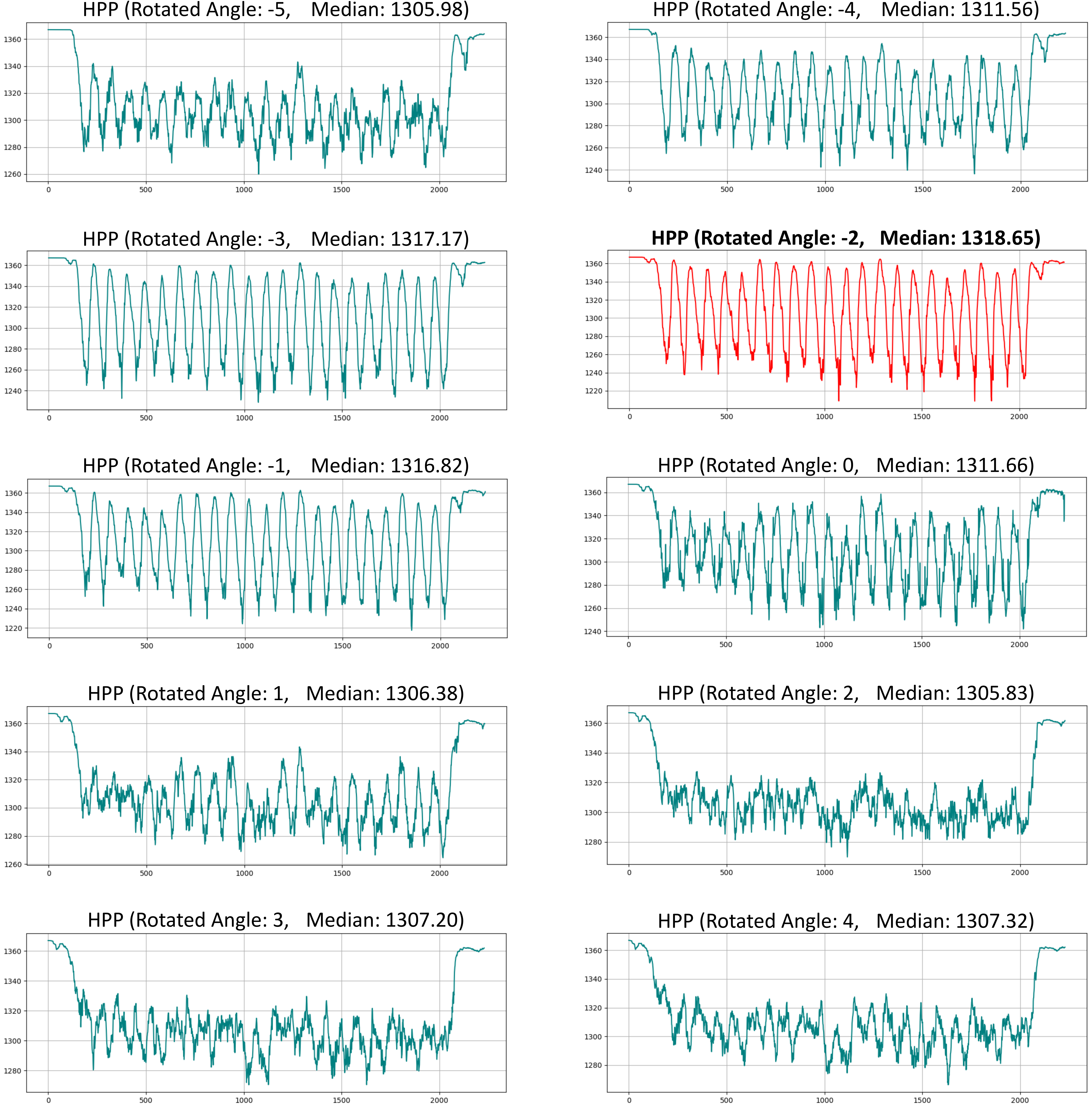}
          \caption{Horizontal Projection Profiles (HPPs) of the Rotated Images; transformed skewed image given in Figure \ref{fig:skewness-to-deskewness}c has been rotated at ten angles ranging from [-5, 5). If you observe, \textit{the HPP of the image rotated at angle '-2' has highest median compared to others}. It has smooth sharp high peaks (a very uniformed distribution of pixel intensities), indicating regions of strong horizontal alignment in the image. It shows that text in skewed image has a more consistent horizontal alignment at this angle.}
      \Description{}\label{fig:rotated-imgs-hpps}
    \end{figure}
   
    \item Subsequently, the median of the HPP of each rotated image is computed, as listed in Table \ref{tbl:pred-angle} which serves as an indicator of alignment. Lower median values indicate that text has poor/less consistent horizontal alignment while higher median values show that text in the image has more consistent horizontal alignment. The HPP of a zero-skewed image will likely have the highest median.

    \begin{table*}[h!]
      \caption{Median of the HPP of the Rotated Images Corresponding to Different Angles}
      \label{tbl:pred-angle}
      
      \begin{tabular}{>{\centering\arraybackslash}p{1cm}>{\centering\arraybackslash}p{5cm}>{\centering\arraybackslash}p{0.9cm}>{\centering\arraybackslash}p{5cm}}
        \toprule
        Angle & Median of HPP (Rotated Image) & Angle & Median of HPP (Rotated Image) \\
        
        \midrule
        -5 & 1305.98 & 0 & 1311.66 \\
        -4 & 1311.56 & 1 & 1306.38 \\
        -3 & 1317.17 & 2 &  1305.83 \\
        \cellcolor{lightblue} \textbf{-2} & \cellcolor{lightblue} \textbf{1318.65} & 3 & 1307.20 \\
        -1 & 1316.82 & 4 & 1307.32 \\
        \bottomrule
      \end{tabular}
      \footnotetext{The optimal rotation angle for deskewing is ‘-2’.}
    \end{table*}
    
    \item Finally, the median of the HPPs is examined and the angle that corresponds to the highest median is selected as the optimal rotation angle for deskewing. In considered example, the determined optimal rotation angle is ‘-2’ as the median of the HPP of the rotated image corresponding to this angle is maximum, as can be seen in Table \ref{tbl:pred-angle}. The skewed image, after being rotated to this determined optimal angle, is depicted in Figure \ref{fig:skewness-to-deskewness}d.

\end{enumerate}

\begin{figure*}[h!]
  \centering     
  \includegraphics[width=12.7cm]{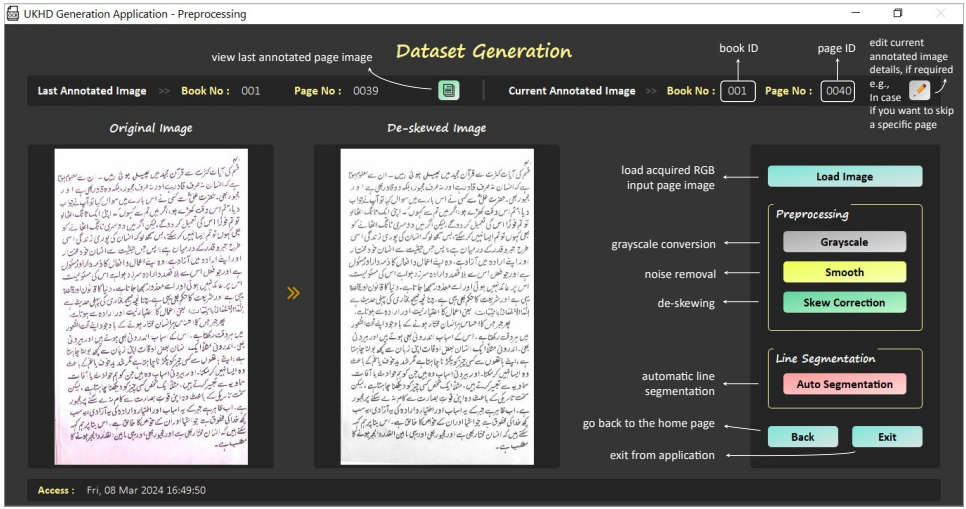}
  \caption{UKHD Generation Application Interface -- Preprocessing Phase} 
  \Description{}
  \label{fig:ukhdg-phase2}
\end{figure*}

All the above steps taken in preprocessing phase were performed on the \textit{\textbf{UKHD Generation Application}}, a desktop application that we have specifically developed for dataset generation. Figure \ref{fig:ukhdg-phase2} shows its interface for the preprocessing phase.

\subsection{Line Segmentation}
In this phase, the preprocessed image was segmented into distinct sub-components i.e. text line images. To achieve this, a semi-automatic approach has been presented that first performs auto-line segmentation based on a horizontal projection profile-based method, and then manually adjusts these auto-segmented text lines if needed. Below is a detailed step-by-step explanation of auto-line segmentation, along with visual representations of an example.

\subsubsection{Binarization}
\begin{figure}[h!]
  \centering       
  \includegraphics[width=6cm]{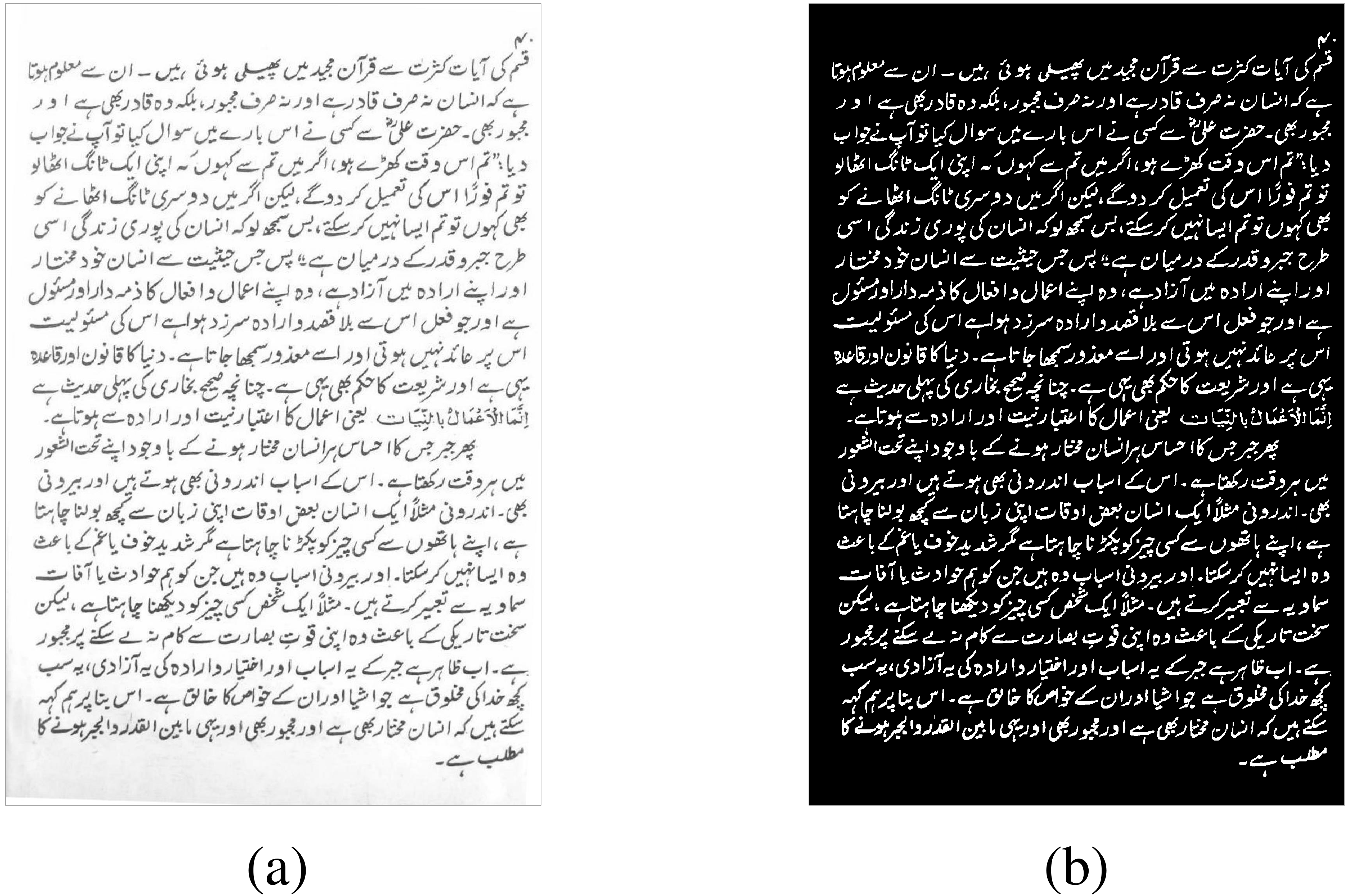}
  \caption{\textbf{(a)} Preprocessed Image, \textbf{(b)} After Applying Otsu's Threshold}
  \Description{}
  \label{fig:binarization}
\end{figure}

Initially, the preprocessed image given in Figure \ref{fig:binarization}a was converted into a binary image using \textit{Otsu’s method}\footnote{\textit{\textbf{Otsu’s} method} is a global thresholding technique which automatically selects an optimal threshold that separates the foreground (text in this case) from the background.} in which white pixels represent the text while black pixels represent the background, as shown in Figure \ref{fig:binarization}b.

\subsubsection{HPP Calculation and Estimating the Potential Regions for Line Boundaries}
\begin{figure}[h!]
  \centering       
  \includegraphics[width=12.7cm]{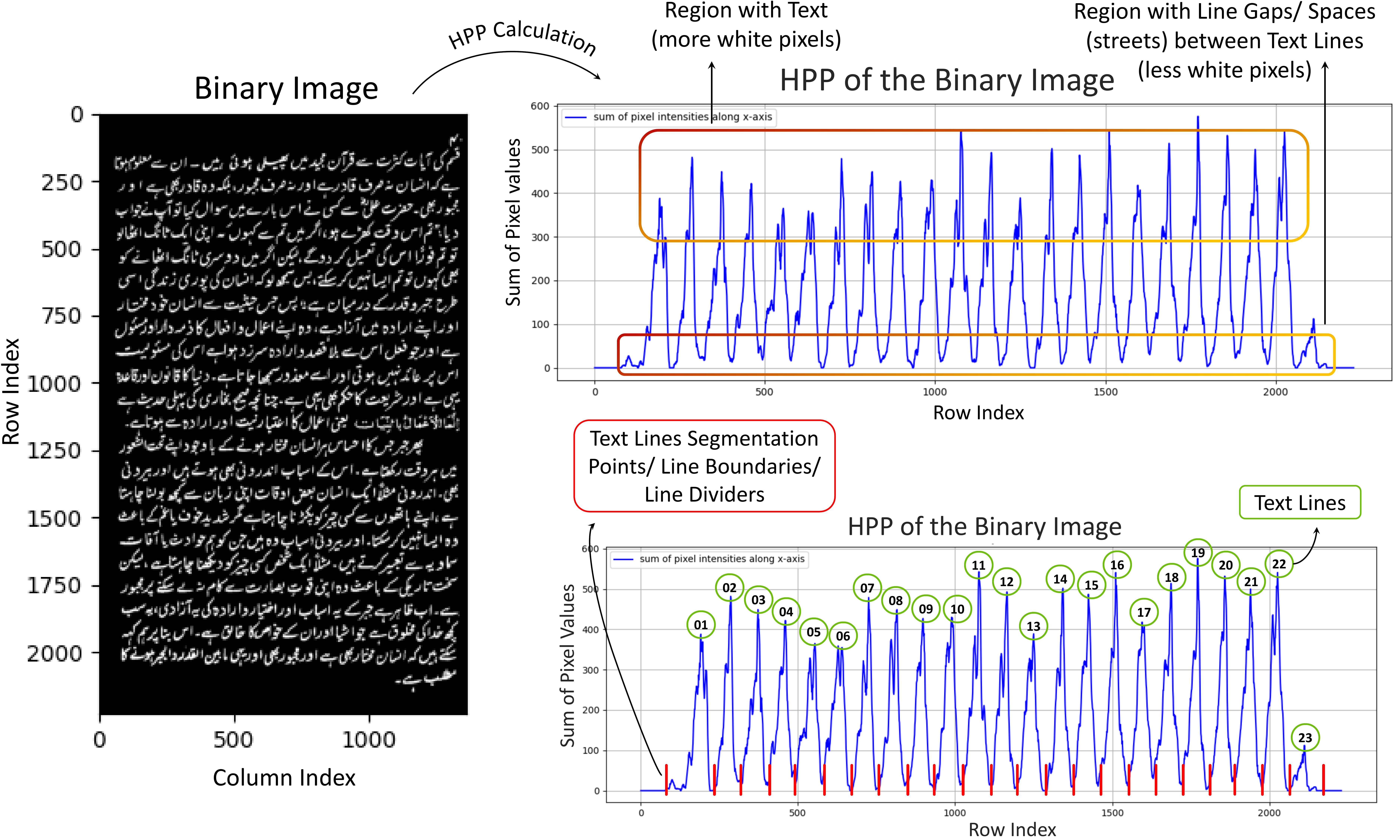}
  \caption{Horizontal Projection Profile (HPP) of the Resultant Binary Image}
  \Description{}
  \label{fig:bw-hpp}
\end{figure}

The horizontal projection of the resultant binary image was then computed i.e. the sum of white pixel values along each row of the binary image, as shown in Figure \ref{fig:bw-hpp}. It helps in examining the distribution of pixel intensities along the horizontal axis of the image i.e. the regions where the text occurrence is high (more white pixels) and the regions where it is less (fewer white pixels), as illustrated in Figure \ref{fig:bw-hpp}.

\begin{figure}[h!]
  \centering       
  \includegraphics[width=12cm]{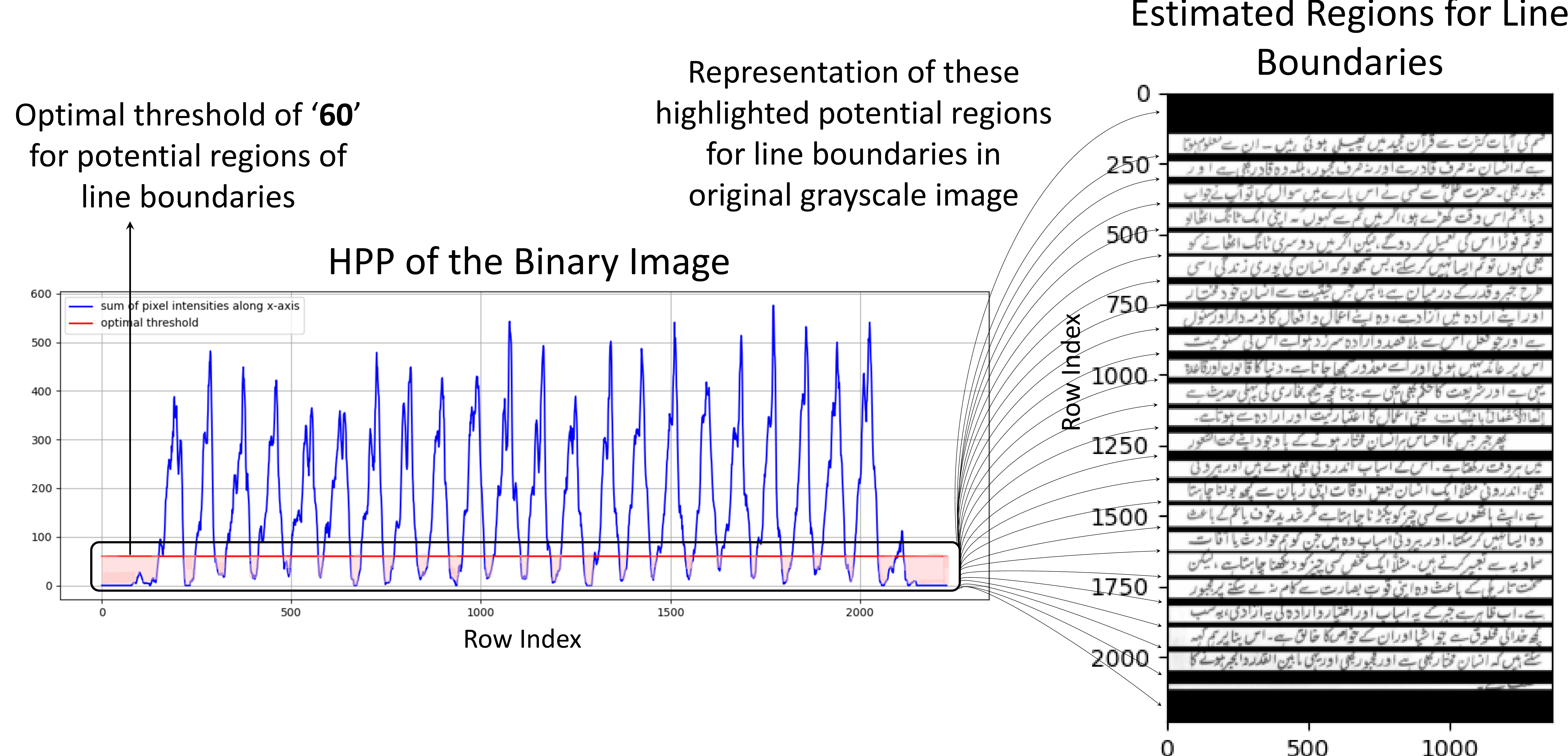}
  \caption{The red line in HPP of the binary image is the optimal threshold for line boundaries. The regions below this threshold are the spaces/gaps between text lines where the line boundaries are located. These estimated regions of line boundaries are also highlighted in input image at right side in which the pixel values of the rows at these indices (estimated regions indexes) are set to ‘0’ (\textit{black regions in the image are estimated regions for line boundaries}).}
  \Description{}
  \label{fig:estimated-line-boundaries}
\end{figure}

The valley (minima) between two successive peaks (maxima) in HPP serves as an indicator for identifying the separation or boundary between two adjacent lines. Therefore, an optimal threshold of ‘\textbf{60}’ has been set for the potential regions of \textit{line boundaries}\footnote{\textit{\textbf{Line boundaries}} are basically the spaces or streets between text lines which separates two adjacent text lines.}. It means the rows of the resultant binary image in which the sum of white pixels is less than or equal to 60 are the estimated potential regions for line boundaries, as depicted in Figure \ref{fig:estimated-line-boundaries}.

\begin{figure}[h!]
  \centering       
  \includegraphics[width=12cm]{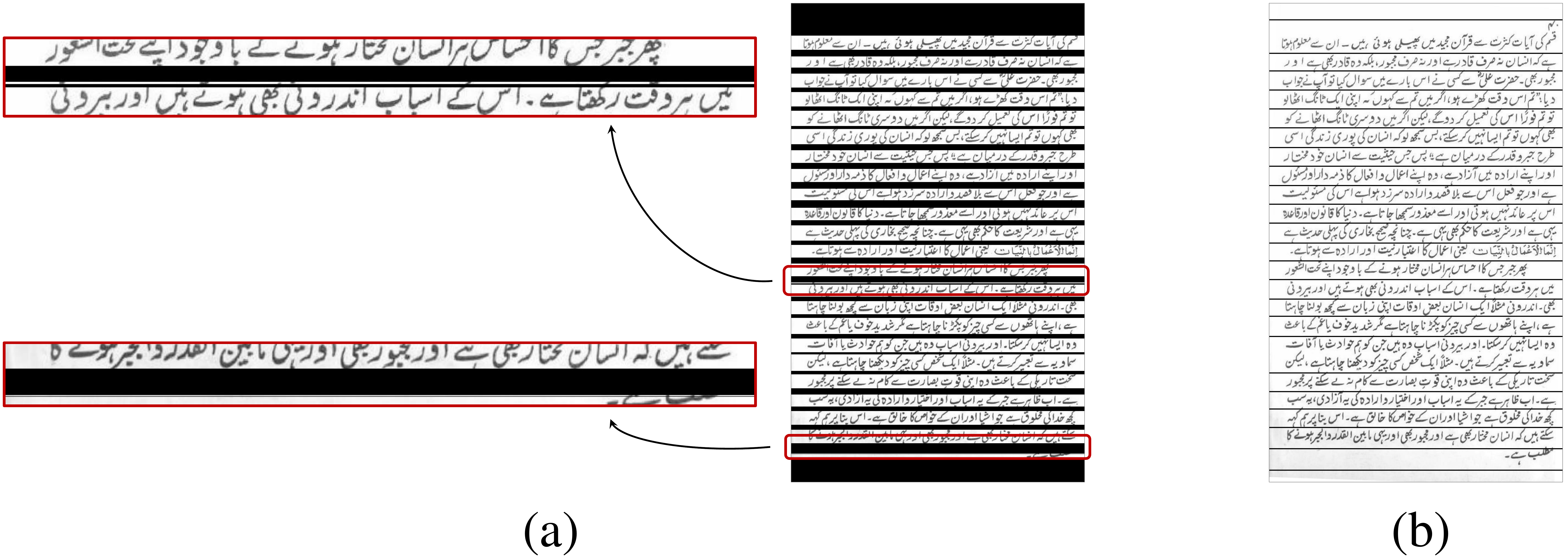}
  \caption{\textbf{(a)} Invalid Estimated Regions of Line Boundaries \textit{(zoom the image and see these regions are causing over-segmentation)}, \textbf{(b)} Detected Line Boundaries}
  \Description{}
  \label{fig:detected-lines}
\end{figure}

The estimated regions for line boundaries were then further refined by excluding those regions that did not actually correspond to a line boundary. This exclusion was necessary because some of these regions are caused by diacritics, as illustrated in Figure \ref{fig:detected-lines}a. To address this, a threshold has been set that examined the height of these estimated regions i.e. it must be greater than \textbf{ten} consecutive rows, estimated region with a height less than threshold has been ignored. This refinement rarely caused \textit{under-segmentation}\footnote{In \textit{\textbf{over-segmentation}}, a single line is miss-segmented into multiple lines while in \textit{\textbf{under-segmentation}}, multiple lines are segmented as a single line.}, however overall it improved auto-line segmentation results by preventing the problem of \textit{over-segmentation}\footnotemark[\value{footnote}].

\subsubsection{Line Boundaries Detection and Text Lines Segmentation}
\begin{figure}[h!]
  \centering       
  \includegraphics[width=9cm]{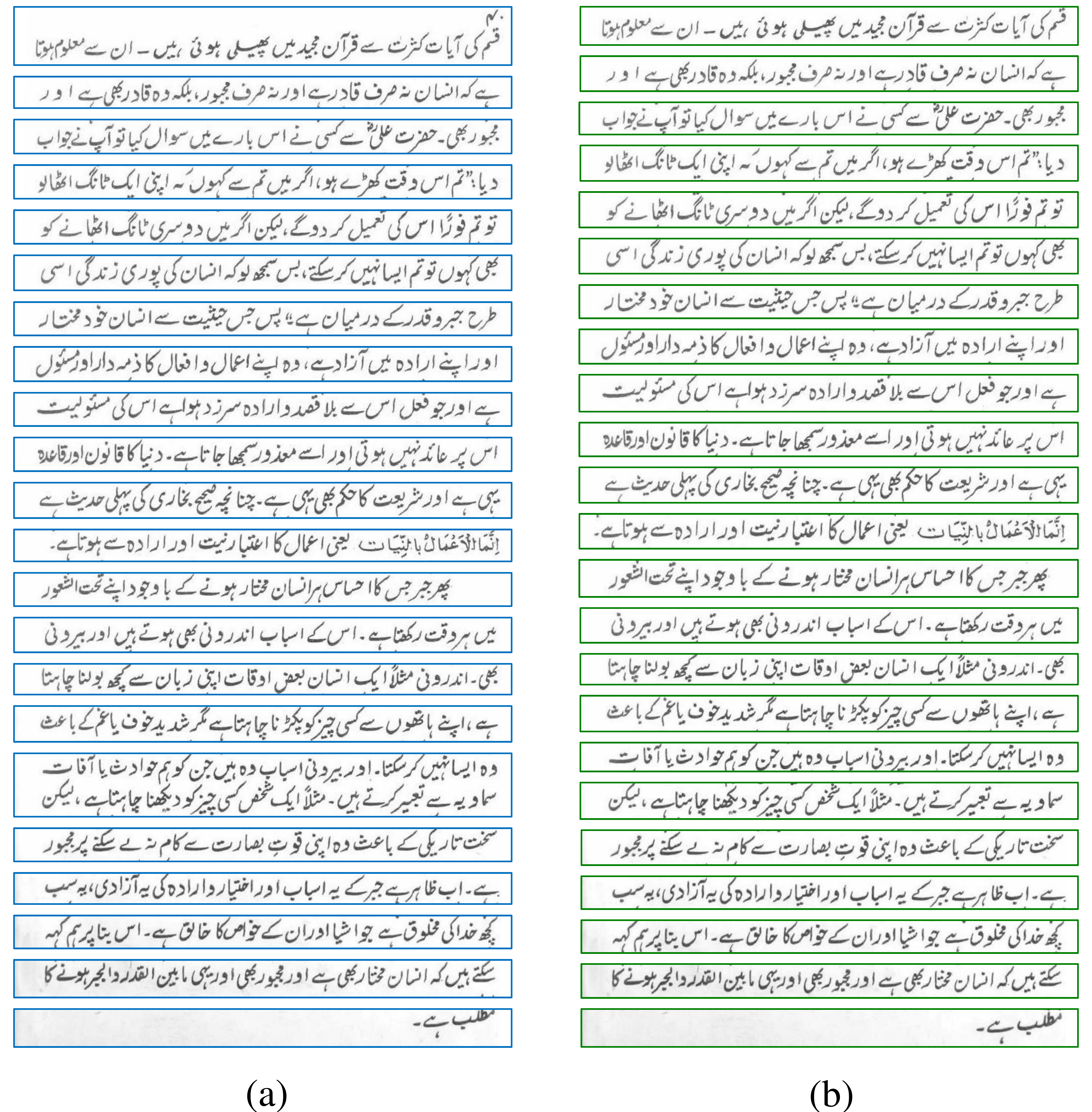}
  \caption{\textbf{(a)} Auto-segmented Text lines, \textbf{(b)} After Manual Adjustments}
  \Description{}
  \label{fig:auto_vs_manual_seg_lines}
\end{figure}

The medians (center indexes) of the refined estimated regions were then computed, indicating the locations of the line boundaries, as shown in Figure \ref{fig:detected-lines}b. These detected line boundaries are the suitable locations for line segmentation, so the lines were cropped out using these locations. These cropped text lines are termed as \textit{auto-segmented text lines}, see Figure \ref{fig:auto_vs_manual_seg_lines}a.

If you observe the auto-segmented text lines in Figure \ref{fig:auto_vs_manual_seg_lines}a which are automatically segmented using above the horizontal projection profile-based method, then you will see that they are not in an optimal form. There are some instances of incorrect segmentation, so these were further adjusted manually. These manual adjustments were carried out within the UKHD generation application, as illustrated in Figure \ref{fig:ukhdg-phase3-4}. Using this application, the positioning of auto-segmented text line images was fine-tuned by shifting them up or down or by cropping them from above or below, as needed. It means, it enabled us to add some area above or below to the text line image from the page image in case of over-segmentation, and crop the text line image from top or bottom in case a line was under-segmented during auto-line segmentation. This application also provided access to retrieve the last annotated text line image to make the necessary adjustments to resolve the under-segmentation problem. Furthermore, it allowed a text line image to be skipped if it deemed unnecessary.

The final segmented text line images after some manual adjustments is depicted in Figure \ref{fig:auto_vs_manual_seg_lines}b, that were subsequently saved in ‘\textbf{.png}’ format with a unique image ID after doing annotation. The image ID format follows the pattern \textit{`bbb-pppp-ll'} having 11 letters including `-' symbol, where \textit{b, p, l} = 1, 2, 3, ... ,9. The first three digits represent the book ID \textit{(e.g. \textbf{001}-pppp-ll)} that categorizes the text lines by their source books. The next four digits denote the page ID \textit{(e.g. 001-\textbf{0040}-ll)} that distinguish between different pages within the same book. The last two digits represent the line ID \textit{(e.g. 001-0040-\textbf{01})} that help to identify the individual text lines extracted from specific page. This image ID uniquely identifies each text line image and its corresponding transcription in UKHD.

\subsection{Annotation}
\label{subsec:annotation}
Manually labeling/transcribing a large amount of data is quite a difficult task as it requires a lot of time, cost and human effort. Therefore, a systematic semi-automatic approach has been implemented for labeling the text line images that combines automated transcription with manual correction.

\subsubsection{Automated Transcription}
Google Cloud Vision offers powerful text recognition capabilities across multiple languages including Urdu. Despite encountering recognition errors in handwritten text, it provides good accuracy. Therefore, the `\textbf{Cloud Vision API}' has been used to transcribe the text from segmented text line images, referred to as \textit{automatic transcription}. This has proven to be very effective in labeling the text line images, as it has resulted in significant time savings.

\subsubsection{Manual Correction}
\begin{figure*}[h!]
  \centering     
  \includegraphics[width=12.7cm]{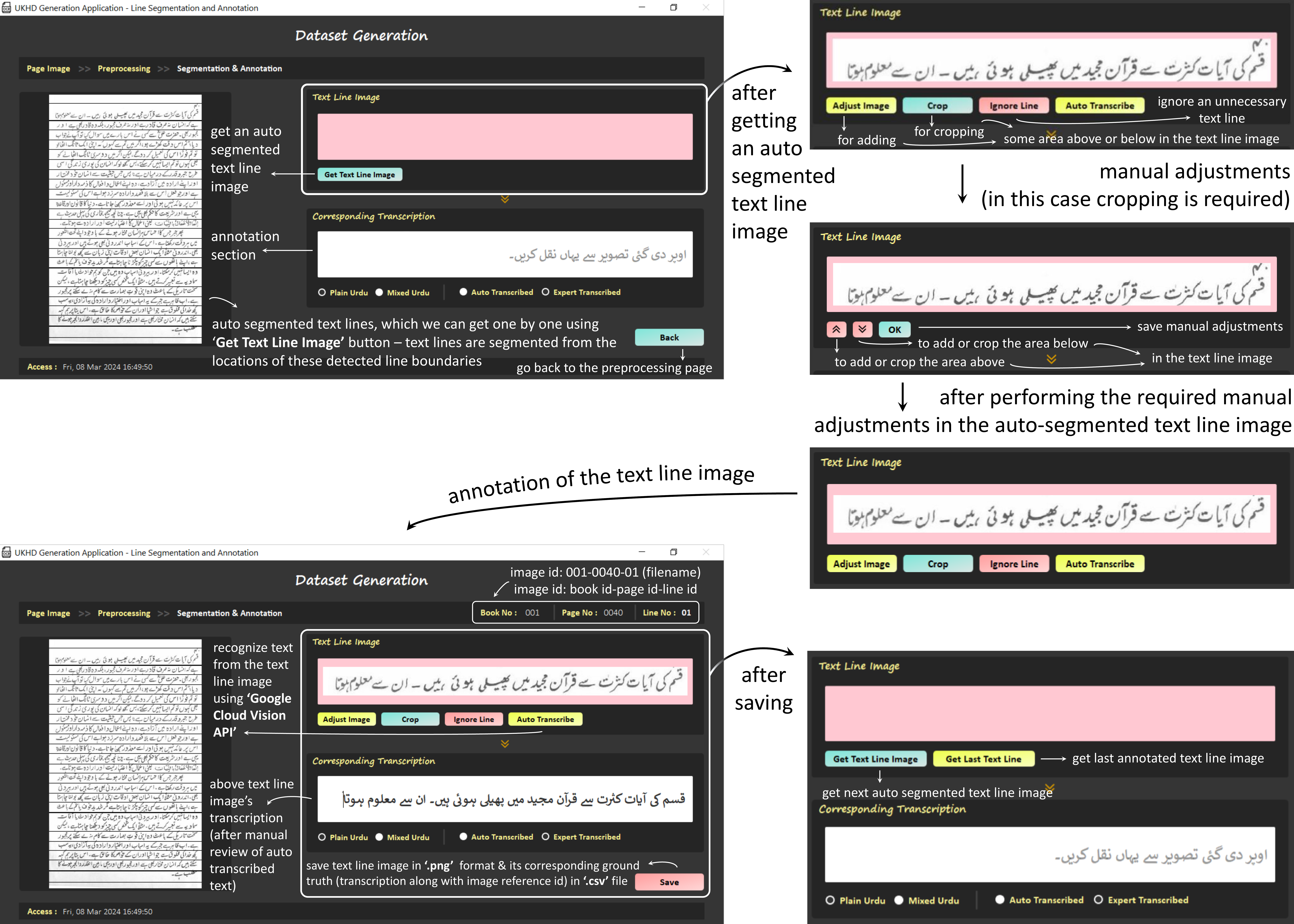}
  \caption{UKHD Generation Application Interface -- Line Segmentation \& Annotation Phase}
  \Description{}
  \label{fig:ukhdg-phase3-4}
\end{figure*}

There were errors in automatic transcription. Therefore, it was further subjected to manual review in which annotator (human expert) reviewed it carefully and corrected the recognition errors. It was also done within the UKHD generation application, as shown in Figure \ref{fig:ukhdg-phase3-4}. The text after manual corrections is the final transcription of the text line image, that was subsequently associated with its corresponding image ID and saved to its respective `\textbf{.csv}' file i.e. PUTL Labels or MUTL Labels. These .csv files serve as structured UKHD ground truth files containing both the transcriptions of text line images and their corresponding references i.e. image IDs.

\section{Implementation of Hybrid Models on UKHD}
\label{sec:implementation}
Four distinct CRNN-based hybrid models, encompassing the CNN-LSTM-CTC model, CNN-BLSTM-CTC model, CNN-GRU-CTC model, and the CNN-BGRU-CTC model, have been analyzed for UKHR.

\subsection{Model Architecture}
\begin{figure*}[ht!]
  \centering     
  \includegraphics[width=12.7cm]{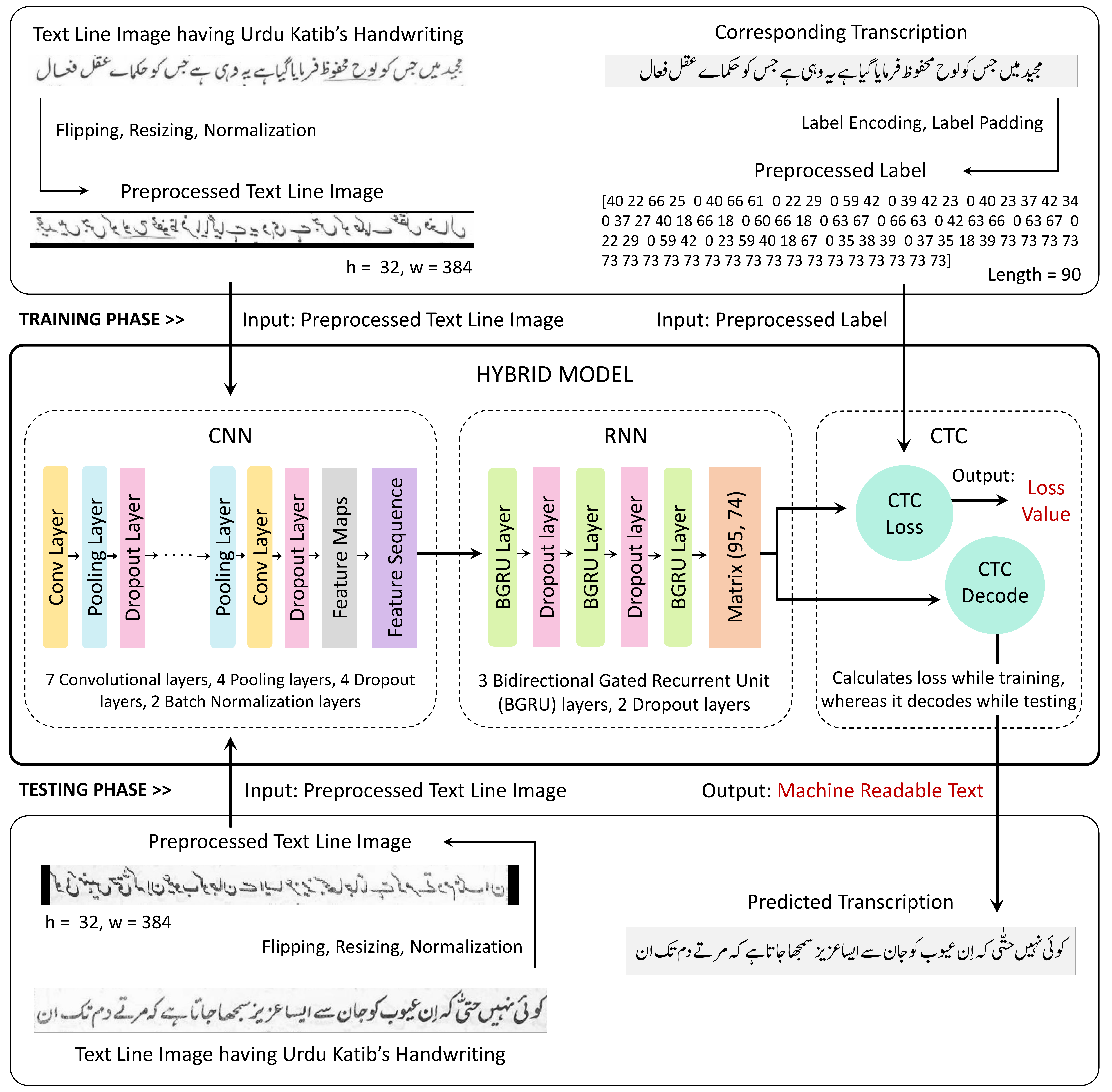}
  \caption{UKHR Model Architecture -- This is essentially the CNN-BGRU-CTC hybrid model architecture, which yielded the best results for UKHR. Therefore, we are calling it UKHR model architecture. Configuration details are provided in Table \ref{tbl:ukhr-model-configurtion}. \textit{(The other three CRNN-based hybrid models have the same architecture except for a slight variation in the RNN component, where BGRU layers are replaced with alternating variants of RNN layers. For example, CNN-LSTM-CTC model replaces BRGU layers with LSTM layers)}}
  \Description{}
  \label{fig:ukhr-model}
\end{figure*}

The model architecture for UKHR in Figure \ref{fig:ukhr-model}, comprises three main parts: CNN, RNN, and CTC. The detailed explanation of each component is elaborated in the following sections.

\begin{table*}[ht]
  \caption{CNN-BGRU-CTC Hybrid Model Configuration -- \textbf{Abbreviations:} 2D Convolutional layer (Conv2D), 2D Max Pooling layer (MaxPool2D), Batch Normalization layer (BN), Bidirectional GRU layer (BGRU) where GRU stands for Gated Recurrent Unit, Connectionist Temporal Classification layer (CTC). In Dense layer configuration, char\_list refers to the number of \textit{classes} (unique of characters and symbols) that is 73, and plus 1 refers to the CTC special blank label. \textit{(Visual representation of the Conv2D layers' output is depicted in Figure \ref{fig:feature-maps})}}
  \label{tbl:ukhr-model-configurtion}
  \centering
  \begin{tabularx}{\textwidth}{p{0.6cm}p{1.5cm}p{7.7cm}X}
    \toprule
    Layer No. & Layer Type & \multirow{2}{*}{Configuration/ Description} & \multirow{2}{*}{Output Shape} \\
    \midrule
    \rowcolor{darkgray!20}
    1 & Input & grayscale text line image, height=32, width=384 & (None, 32, 384, 1) \\
    
    \addlinespace
    \rowcolor{Goldenrod!35}
    2 & Conv2D & filters=64, kernel=3x3, activation=relu, padding=same & (None, 32, 384, 64) \\
    
    \rowcolor{SkyBlue!20}
    3 & MaxPool2D & pool size=2x2 & (None, 16, 192, 64) \\
    
    \rowcolor{CarnationPink!25}
    4 & Dropout & dropout rate=0.3 & (None, 16, 192, 64) \\
    
    \rowcolor{Goldenrod!35}
    5 & Conv2D & filters=128, kernel=3x3, activation=relu, padding=same & (None, 16, 192, 128) \\
    
    \rowcolor{SkyBlue!20}
    6 & MaxPool2D & pool size=2x2 & (None, 8, 96, 128) \\
    
    \rowcolor{CarnationPink!25}
    7 & Dropout & dropout rate=0.3 & (None, 8, 96, 128) \\
    
    \rowcolor{Goldenrod!35}
    8 & Conv2D & filters=256, kernel=3x3, activation=relu, padding=same & (None, 8, 96, 256) \\
    
    \rowcolor{Goldenrod!35}
    9 & Conv2D & filters=256, kernel=3x3, activation=relu, padding=same & (None, 8, 96, 256) \\
   
    \rowcolor{SkyBlue!20} 
    10 & MaxPool2D & pool size=2x1 & (None, 4, 96, 256) \\
    
    \rowcolor{Goldenrod!35}
    11 & Conv2D & filters=512, kernel=3x3, activation=relu, padding=same & (None, 4, 96, 512) \\
    
    \rowcolor{CarnationPink!25}
    12 & Dropout & dropout rate=0.3 & (None, 4, 96, 512) \\
    
    \rowcolor{Brown!15}
    13 & BN & --- & (None, 4, 96, 512) \\
    
    \rowcolor{Goldenrod!35}
    14 & Conv2D & filters=512, kernel=3x3,  activation=relu, padding=same & (None, 4, 96, 512) \\
    
    \rowcolor{Brown!15}
    15 & BN & --- & (None, 4, 96, 512) \\
    
    \rowcolor{SkyBlue!20}
    16 & MaxPool2D & pool size=2x1 &	(None, 2, 96, 512) \\
    
    \rowcolor{Goldenrod!35}
    17 & Conv2D & filters=512, kernel=2x2, activation=relu, padding=same & (None, 1, 95, 512) \\
    
    \rowcolor{CarnationPink!25}
    18 & Dropout & dropout rate=0.3 & (None, 1, 95, 512) \\
    
    \rowcolor{Periwinkle!30}
    19 & Lambda & squeeze along axis 1 (removes singleton dimension) & (None, 95, 512) \\

    \addlinespace
    \rowcolor{SpringGreen!50}
    20 & BGRU & hidden units=512, return sequences=true & (None, 95, 1024) \\
    
    \rowcolor{CarnationPink!25}
    21 & Dropout & dropout rate=0.3 & (None, 95, 1024) \\
    
    \rowcolor{SpringGreen!50}
    22 & BGRU &	hidden units=512, return sequences=true & (None, 95, 1024) \\
    
    \rowcolor{CarnationPink!25}
    23 & Dropout & dropout rate=0.3 & (None, 95, 1024) \\

    \rowcolor{SpringGreen!50}
    24 & BGRU & hidden units=512, return sequences=true & (None, 95, 1024) \\
    
    \rowcolor{Apricot!50}
    25 & Dense & no. of units=len(char\_list)+1, activation=softmax & (None, 95, 74) \\

    \addlinespace
    \rowcolor{JungleGreen!30}
    26 & CTC & loss calculation and decoding & Output \\
    \bottomrule
  \end{tabularx}
\end{table*}

\subsubsection{CNN}
Convolutional Neural Networks (CNNs) are widely used for computer vision tasks due to their ability to learn patterns from images. A typical CNN architecture consists of three types of layers: “convolutional layers”, “pooling layers”, and “fully connected layer”. Only convolutional and pooling layers have been utilized in the designed models for UKHR. In \textbf{convolutional} layers, a small matrix known as \textit{filter}\footnote{ The \textit{\textbf{filters}} are basically a set of learnable parameters also called \textit{feature detectors} as they detect certain features in the image such as edges.} or \textit{kernel}, is convolved over the input image to extract the local and abstract patterns. Whereas \textbf{pooling} layers are responsible for down-sampling, they reduce the dimensionality of the \textit{feature maps} generated after the convolution operation while preserving the pertinent features, which facilitate subsequent processing i.e. decreasing the number of computations. Further in-depth details about the CNN can be seen in \cite{albawi2017understanding,o2015introduction}.

\begin{figure*}[h!]
  \centering     
  \includegraphics[width=12.2cm]{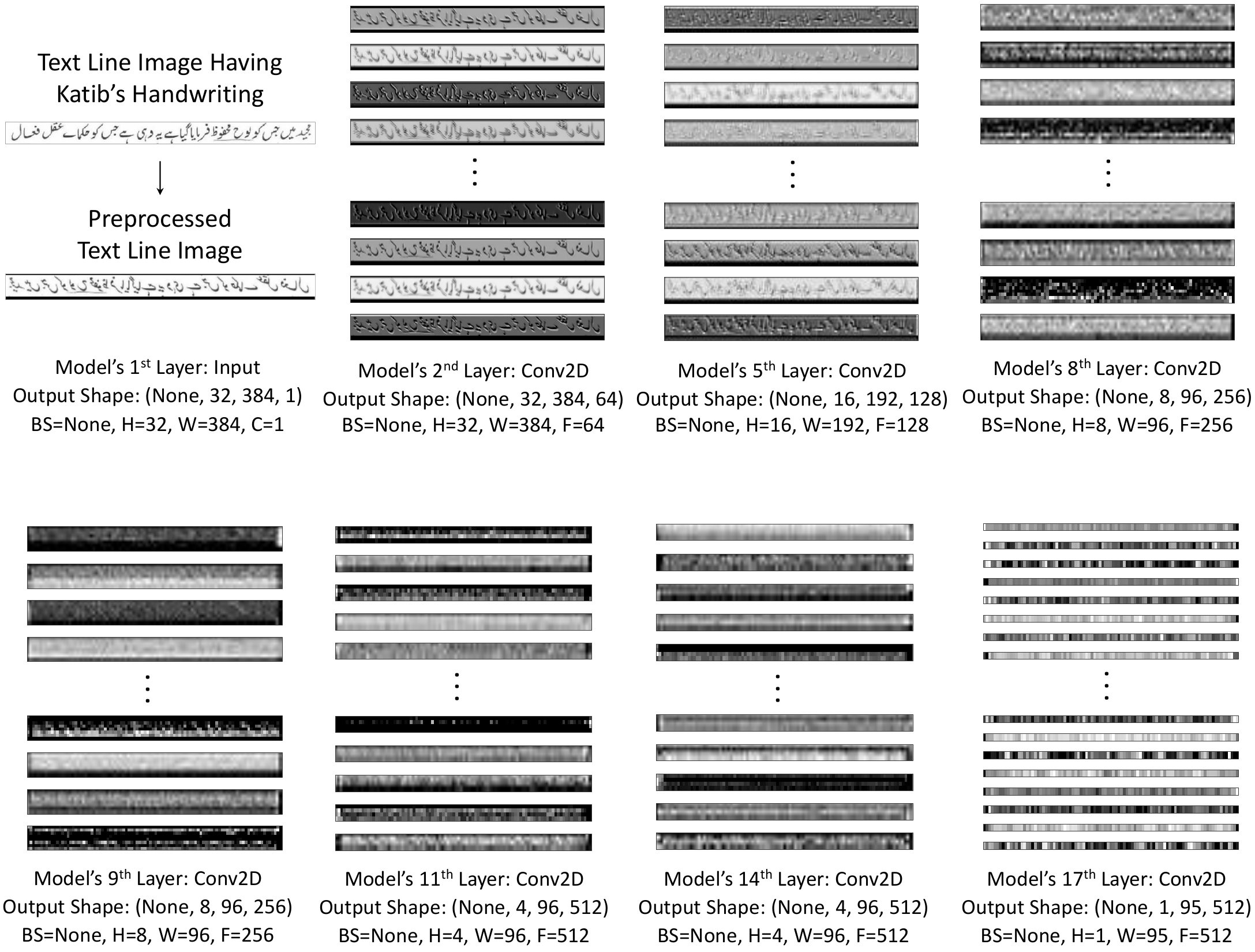}
  \caption{Visualization of feature maps produced by Conv2D layers in the CNN-BGRU-CTC model. The initial convolutional layers extracted the low-level features whereas final convolutional layers extracted the high-level features. \textbf{Abbreviations:} Channels (C), Batch Size (BS), Height (H), Width (W), Filters (F). Technically `F' denotes filters; however, for easy understanding, it can be referred to as feature maps. It represents the number of distinct features or patterns captured during convolutional operations.}
  \Description{}
  \label{fig:feature-maps}
\end{figure*}

The CNN component of the UKHR model in Figure \ref{fig:ukhr-model}, serves as the \textit{feature extractor} part in which several convolutional layers are stacked, possibly followed or not by a max pooling layer, dropout layer, or a batch normalization layer. The configuration of these layers are given in Table \ref{tbl:ukhr-model-configurtion}; all the layers up to the lambda layer are composing the CNN component. This part of model is responsible for capturing hierarchical and spatial features from the input image. It processes a grayscale text line image with dimensions of 32x384. In initial layers, it extracts the local features corresponding to basic visual elements found in text such as edges, strokes, corners, variations in stroke thickness etc. As the network becomes deeper, it begins to capture more complex and abstract patterns, structures and segmentation points in the text. Figure \ref{fig:feature-maps} shows the resulting feature maps, highlighting both local and global features of the input image.

\begin{figure*}[h!]
  \centering    
  \includegraphics[width=12.7cm]{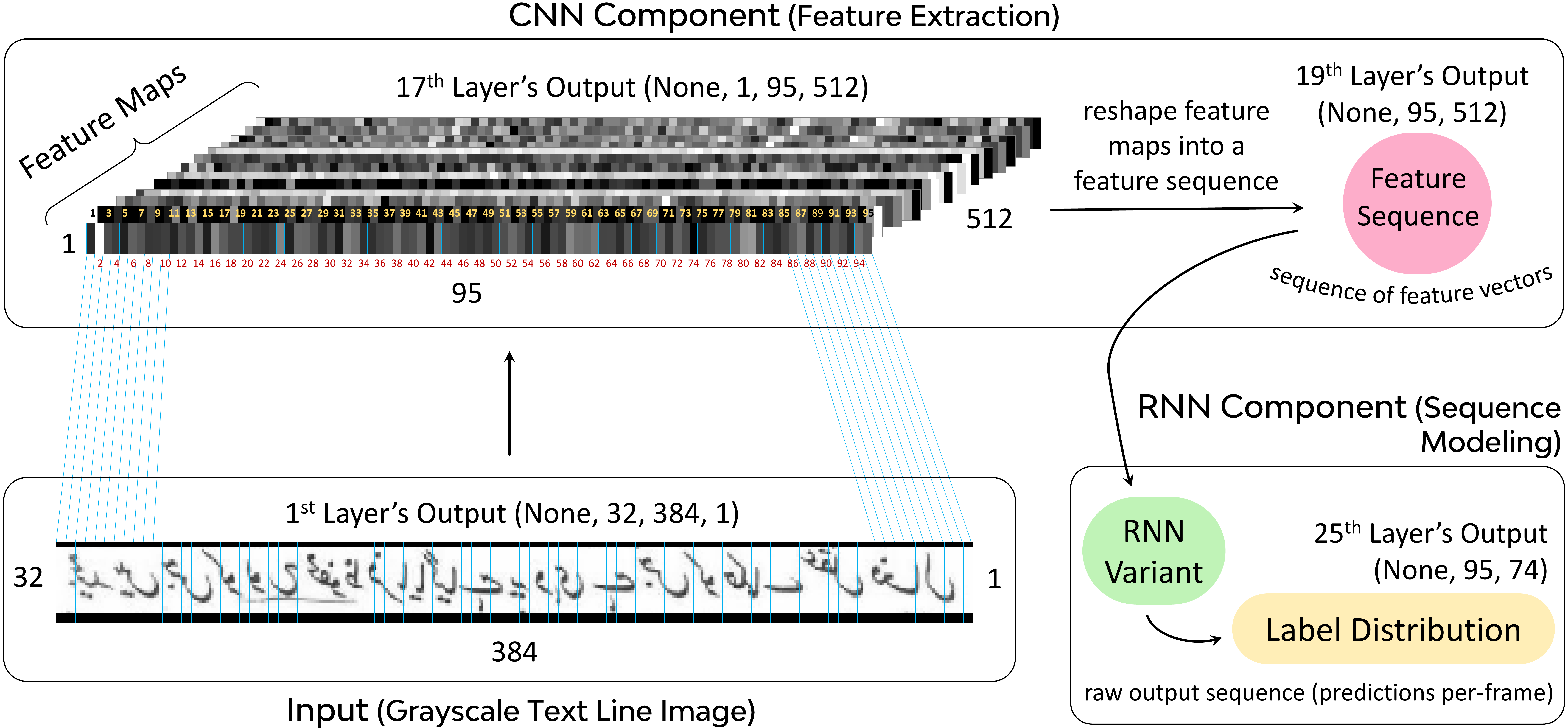}
  \caption{The CNN component takes a grayscale text line image as input and returns a feature sequence. The RNN component then processes this resulting sequence and outputs a label distribution \textit{(it transforms the  feature sequence into per-frame predictions i.e. a raw output sequence)}. \textbf{Dimensions \& Description:} Input Image = 32x384x1 \textit{(height: 32, width: 384, depth/channels: 1)}, Feature Sequence = 95x512 \textit{(no. of time steps: 95, no. of features per time step: 512)}, Label Distribution = 95x74 \textit{(sequence length: 95, classes: 74)}}
  \Description{}
  \label{fig:cnn-rnn-output}
\end{figure*}

Collectively these feature maps are the representation of input text line image features in spatial grid (see Figure \ref{fig:cnn-rnn-output}), which are subsequently converted into a \textit{feature sequence} (sequence of feature vectors). The resulting feature sequence has dimensions of (95x512), where 95 denotes number of time steps, and 512 represents the number of features at each time step. Each time step corresponds to a frame (blue rectangle) in the feature maps, see Figure \ref{fig:cnn-rnn-output}, the width of feature maps is 95 pixels, where each pixel represents one time step. This means there are total 95 time steps in the sequence, and each time step is represented by a feature vector of dimension 512. These feature vectors represent the features extracted from the input text line image at each step along the sequence. This feature sequence serves as an input for the RNN layers.

\subsubsection{RNN}
Recurrent Neural Networks (RNNs) are well-suited for handling sequential data where the sequence of data elements hold significance. Text can be considered as sequential data because it is essentially a sequence of characters. So, in the context of designed hybrid model for UKHR, following the CNN layers, RNN layers are integrated to capture the sequential dependencies between the extracted features.

The conventional RNNs which are designed with self-connected hidden layers, encountered a challenge known as \textit{Vanishing Gradient Problem}, limiting their ability to capture long term dependencies within data. Hochreiter and Schmidhuber introduced the \textbf{Long Short-Term Memory} (LSTM) network \cite{hochreiter1997long}, which replaced the conventional RNN hidden layer's memory cell with a specialized memory cell and three essential gates. This design empowers LSTMs to excel in capturing and preserving long-term dependencies, also making it particularly valuable when dealing with sequences derived from image-based data \cite{shi2016end-to-end}. Likewise in 2014, another network known as the \textbf{Gated Recurrent Unit} (GRU) was introduced by Cho et al.  \cite{cho2014learning}. It is similar to LSTM but has a simpler architecture utilizing only two gates; nonetheless  it offers comparable performance to LSTMs \cite{chung2014empirical,chen2017multi}. Additionally, bidirectional variants of these architectures i.e. \textbf{Bidirectional LSTM} (BLSTM), and \textbf{Bidirectional GRU} (BGRU), provide a more comprehensive representation of data by processing the sequence bidirectionally. Further details regarding the RNN variants can be seen in \cite{hochreiter1997long,chung2014empirical,cho2014learning,staudemeyer2019understanding-lstm}.

In this study, we conducted a systematic evaluation of these RNN variants within a hybrid model framework to assess their effectiveness in sequence modeling, specifically focusing on their performance in UKHR. The adopted approach resulted in the creation of four CRNN-based hybrid models mentioned at the beginning of Section \ref{sec:implementation}. All these models have same architecture except for a slight variation in the RNN component as mentioned in the caption of Figure \ref{fig:ukhr-model}. This figure essentially demonstrates the architecture of the CNN-BGRU-CTC model, called the \textit{UKHR model}, which is being elaborated.

The RNN component in Figure \ref{fig:ukhr-model} of UKHR model consists of three Bidirectional GRU layers, possibly followed or not by a dropout layer. The configuration of these layers is given in Table \ref{tbl:ukhr-model-configurtion}; all the layers after Lambda layer except the CTC layer are composing the RNN component. This part of the model is responsible for \textit{sequence modeling}, it captures the sequential dependencies and contextual relationships among characters and words within the sequence. After sequentially processing the feature sequence obtained from the CNN component (see Figure \ref{fig:cnn-rnn-output}), it returns a \textit{label distribution} \textemdash a probability distribution over all possible classes at each step in the sequence. The resulting distribution, with dimensions of (95x74), can be referred to as a \textit{raw output sequence} containing character predictions (probabilities) at each frame/timestep. Here, 95 indicates the sequence length, and 74 refers to the number of \textit{classes} (the vocabulary size, including an additional CTC blank label). This demonstrates the model's capability to recognize sequences with a maximum length of 95 characters.

\subsubsection{CTC}
\begin{figure*}[h!]
  \centering     
  \includegraphics[width=12.1cm]{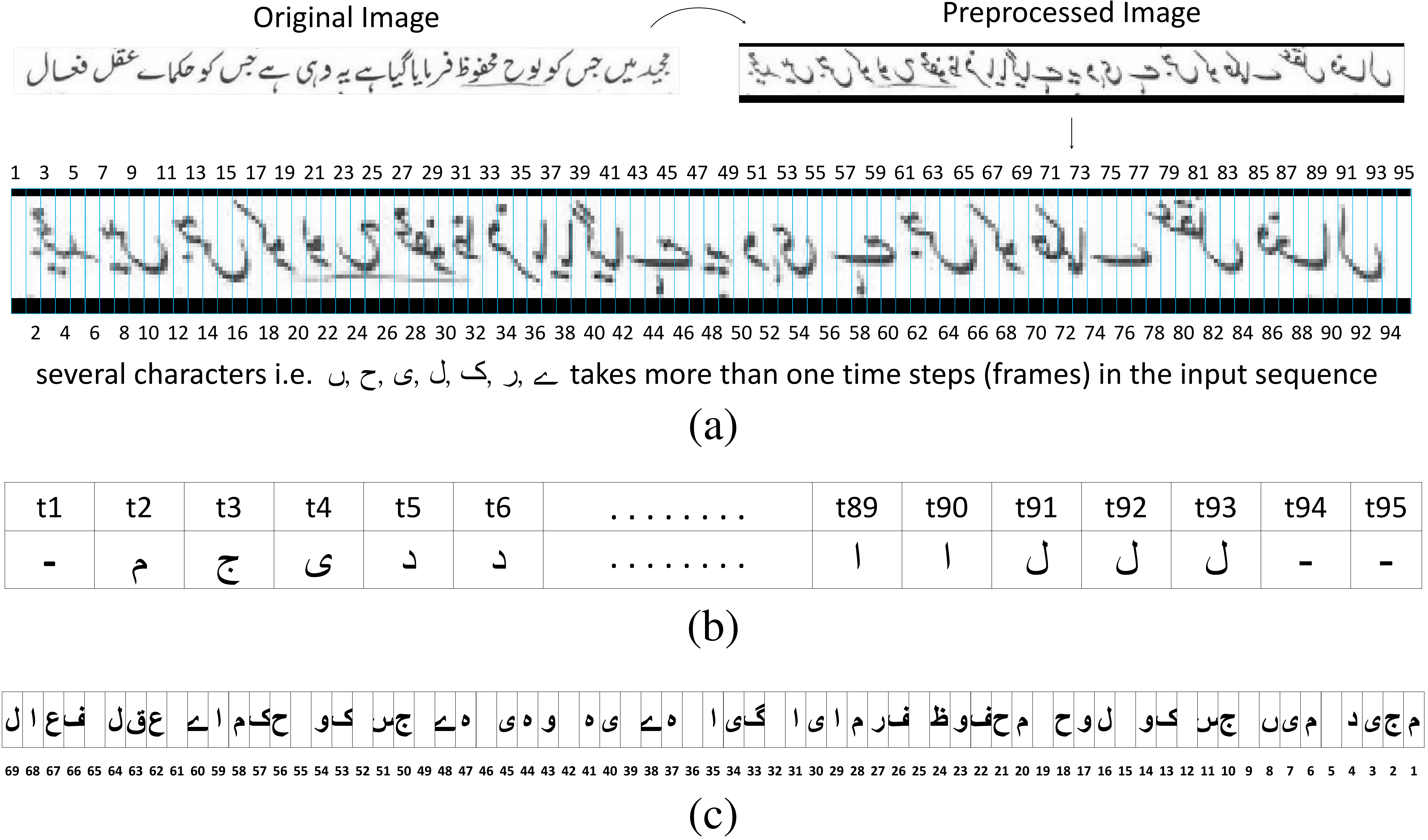}
  \caption{\textbf{(a) Input sequence length is 95}. It is the input of the model (RNNs part input) which consists of a sequence of 95 time steps, where each time step represents a segment of an image. \textbf{(b) Raw output sequence length is 95}. It represents the predictions made by the model (RNNs part output) at each time step of the input sequence. In actual it predicts a probability distribution over all possible characters, including a special CTC blank label. Here for clarity and simplicity, only the highest probability classes (characters) are focused. \textbf{(c) Output sequence or ground truth length is 69}. This is the actual text (transcription) corresponding to the handwriting in the input sequence (image) that the model must learn to predict.}
  \Description{}
  \label{fig:ctc-need}
\end{figure*}

The Connectionist Temporal Classification (CTC) algorithm is designed to handle the scenarios where the alignment between the input sequence and the output sequence is not one-to-one. Like in handwriting recognition, the input and output sequences have variable lengths, see Figure \ref{fig:ctc-need}. Furthermore, in handwriting recognition, the exact alignment between the input (image) and the output (transcription) is unknown. It can be seen in Figure \ref{fig:ctc-need}, several characters in the input image takes more than one time steps, and we do not know which regions of the image aligns to which character in the ground truth. CTC addresses all these problems as it is an alignment-free algorithm. It introduces a special \textit{blank label}, denoted by ‘-’ or ‘$\epsilon$’, which is inserted between the characters of the output sequence to indicate that there is no label at that position (time step) \textemdash also used for handling duplicated/repeated characters. The in-depth details that how CTC works, and its mathematical terms can be seen in \cite{graves2006connectionist}; additionally, to understand how it is used with CRNN-based model for HTR, you can explore these studies \cite{jiang2018baidu,chen2018improvement,tong2020ma,gader2022attention-arabic}.

A CTC layer is added following the RNN component of the UKHR model, see Figure \ref{fig:ukhr-model}. This layer holds a significant role in converting text into a machine-readable format \textemdash often referred to as the \textit{transcription layer}. Here the CTC component has two fundamental functions: loss calculation and decoding the input sequence to the final text sequence. During training, the CTC component receives the label distribution generated by the RNN component along with the ground truth sequence. It computes the \textit{loss} by summing the probabilities of all possible alignments between them. This process allows the model to learn the correct sequence alignment and improve its predictions. While testing, the CTC component only receives the label distribution generated by the RNN component and performs \textit{decoding}. It first determines the best path by identifying the character with the maximum probability at each time step. Then, it merges repeated characters and removes the blank labels to form the final text sequence.

\begin{table}[h!]
  \centering
  \caption{Specifications and Distribution of the Experimental Data  \textemdash \textit{The histogram of labels length and the frequency of each class in dataset can be seen in Figure \ref{fig:putl-hist} and \ref{fig:classes-freq} respectively}.}
  \label{tbl:experimental-data-statistics}
  \begin{tabular}{p{5cm}p{7cm}}
    \toprule
    Description & Statistics\\
    \midrule
    Dataset & PUTL \hspace{0.1cm}\textcolor{teal}{\textit{// primary subset of UKHD}}\\
    Total Images & 11,742 images\\
    Training Set & 9,393 images \hspace{0.2cm} \textcolor{teal}{\textit{// 80\% of the total imgs}}\\
    Validation Set & 1,174 images \hspace{0.2cm} \textcolor{teal}{\textit{// 10\% of the total imgs}}\\
    Testing Set & 1,175 images \hspace{0.2cm} \textcolor{teal}{\textit{// 10\% of the total imgs}}\\
    Maximum Label Length & 90 \hspace{0.1cm} \textcolor{teal}{\textit{// maximum number of chars in a text line}}\\
    No. of Classes/ Vocabulary size & 73 \hspace{0.05cm} \textcolor{teal}{\textit{// unique no. of characters \& symbols in PUTL}}\\
    
    \bottomrule
  \end{tabular}
\end{table}

\begin{figure*}[h]
  \centering     
  \includegraphics[width=12cm]{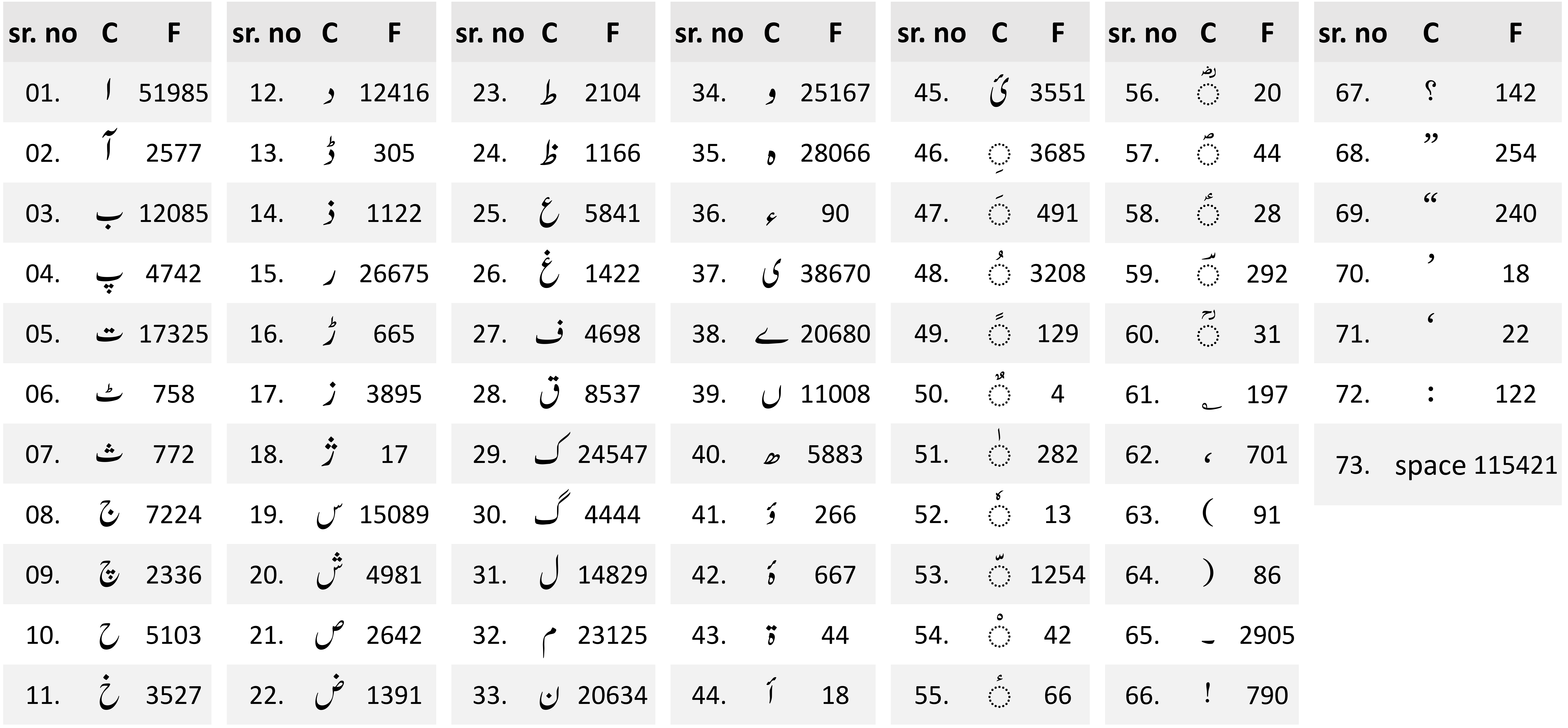}
  \caption{Frequency of Each Class i.e. Character and Symbol, in Experimental Dataset (\textit{PUTL subset}); \textbf{Abbreviations:} Class (C), Frequency (F)}
  \Description{}
  \label{fig:classes-freq}
\end{figure*}

\subsection{Experimental Data}
The Plain Urdu Text Lines (PUTL), primary subset of Urdu Katib Handwritten Dataset (UKHD) having purely Urdu language, has been used for training and evaluation of models designed for UKHR. To ensure robust evaluation, PUTL was partitioned into three distinct subsets: \textit{training set} for model training, \textit{validation set} for fine-tuning hyperparameters, and \textit{test set} for comprehensive performance assessment. The distribution of data as well as some other specifications of PUTL is provided in Table \ref{tbl:experimental-data-statistics}. The PUTL subset has 73 unique characters and symbols, each representing a distinct class for classification purposes, see Figure \ref{fig:classes-freq}.

\subsection{Data Preparation}
This section provides insight into the data preparation process, which includes several steps that ensure the data is in an optimal format that is compatible with the model architecture.

\subsubsection{Preprocessing Text Line Images}
During the creation of UKHD, image preprocessing was done to improve the image quality. Now, this further preprocessing has been performed to align them with the model architecture.

\begin{figure}[ht!]
  \centering     
  \includegraphics[width=12.2cm]{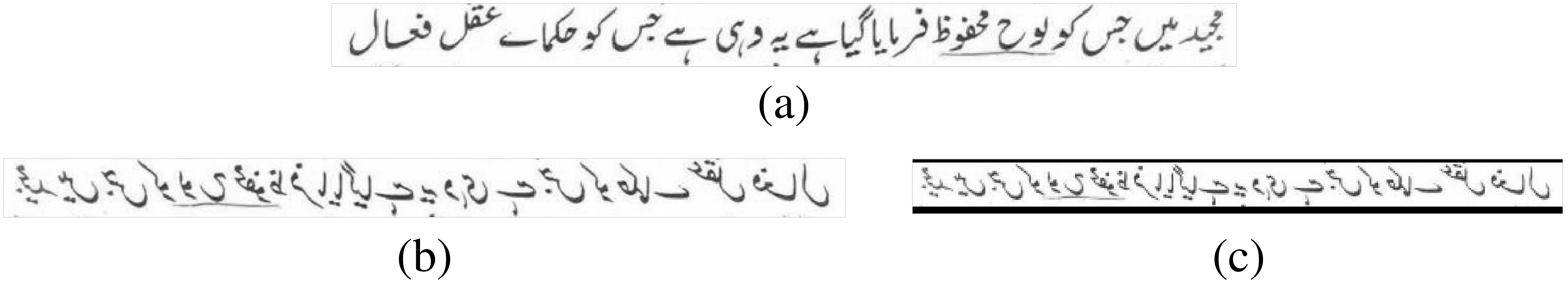}
  \caption{\textbf{(a)} Input Text Line Image: Text is written from right-to-left, \textbf{(b)} After Horizontal Flipping: Reverses the character order in the image (making text left-to-right), \textbf{(c)} After Resizing: After experimenting with different sizes, following standard height and width dimensions were chosen because they maintained the image quality without losing the textual information; \textit{height=32 pixels, width=384 pixels, padding=true, maintain\_aspect\_ratio=true}.}
  \Description{}
  \label{fig:preprocessing-text-line-image}
\end{figure}

\begin{itemize}
    \item[$\bullet$] \textbf{Flipping:} The CRNN-based models typically expect text in a left-to-right format. Their internal processing assumes the character sequence starts from the left and progresses to the right. While Urdu script has the opposite order, see Figure \ref{fig:preprocessing-text-line-image}a. So, the image was flipped horizontally which effectively reverses the character order in the image as shown in Figure \ref{fig:preprocessing-text-line-image}b, making it left-to-right and aligning it with the model's internal processing.
    
    \item[$\bullet$] \textbf{Resizing:} The text line images within UKHD has different dimensions. They have no fixed height and width, which poses complexities for the model to process them. Therefore, all text line images were resized into a standard size i.e. 32 pixels height and 384 pixels width while preserving the aspect ratio; resized image sample is given in Figure \ref{fig:preprocessing-text-line-image}c. This resizing process, inclusive of padding where necessary maintained the structural shape of the text which is important for the accurate recognition.

    \item[$\bullet$] \textbf{Intensity Normalization:} It is a process of normalizing pixel values within an image to a consistent range/scale that enhances model stability, reduces overfitting, and improves generalization. So, the pixel values were re-scaled from their original range of 0 to 255 into a new range of 0 to 1. It was done by dividing each pixel value by 255.
    
\end{itemize}

\subsubsection{Preprocessing Labels}
The designed models operate exclusively with numerical inputs, therefore the following two steps were performed to transform the labels.

\begin{itemize}
    \item[$\bullet$] \textbf{Label Encoding:} It converted the text line image transcriptions (labels) into numerical representations using a character-to-number mapping scheme. Each unique character and symbol in the dataset was assigned a distinct numeric identifier ranging from 1 to 73. In experimental dataset, there are 73 unique characters and symbols, can be seen in Figure \ref{fig:classes-freq}, which indicate the vocabulary size.
    
    \item[$\bullet$] \textbf{Sequence Padding:} Along with label encoding, sequence padding was performed to standardize the length of all labels, as model requires uniform input length. The maximum label length in experimental dataset is 90. Hence, all encoded sequences were padded to this length by appending the number 73 \textit{(the vocabulary size)} to the end of the sequence. This ensures that the padded values do not introduce new characters or symbols and remain consistent with the existing vocabulary.
\end{itemize}

\begin{figure}[ht!]
  \centering     
  \includegraphics[width=12.7cm]{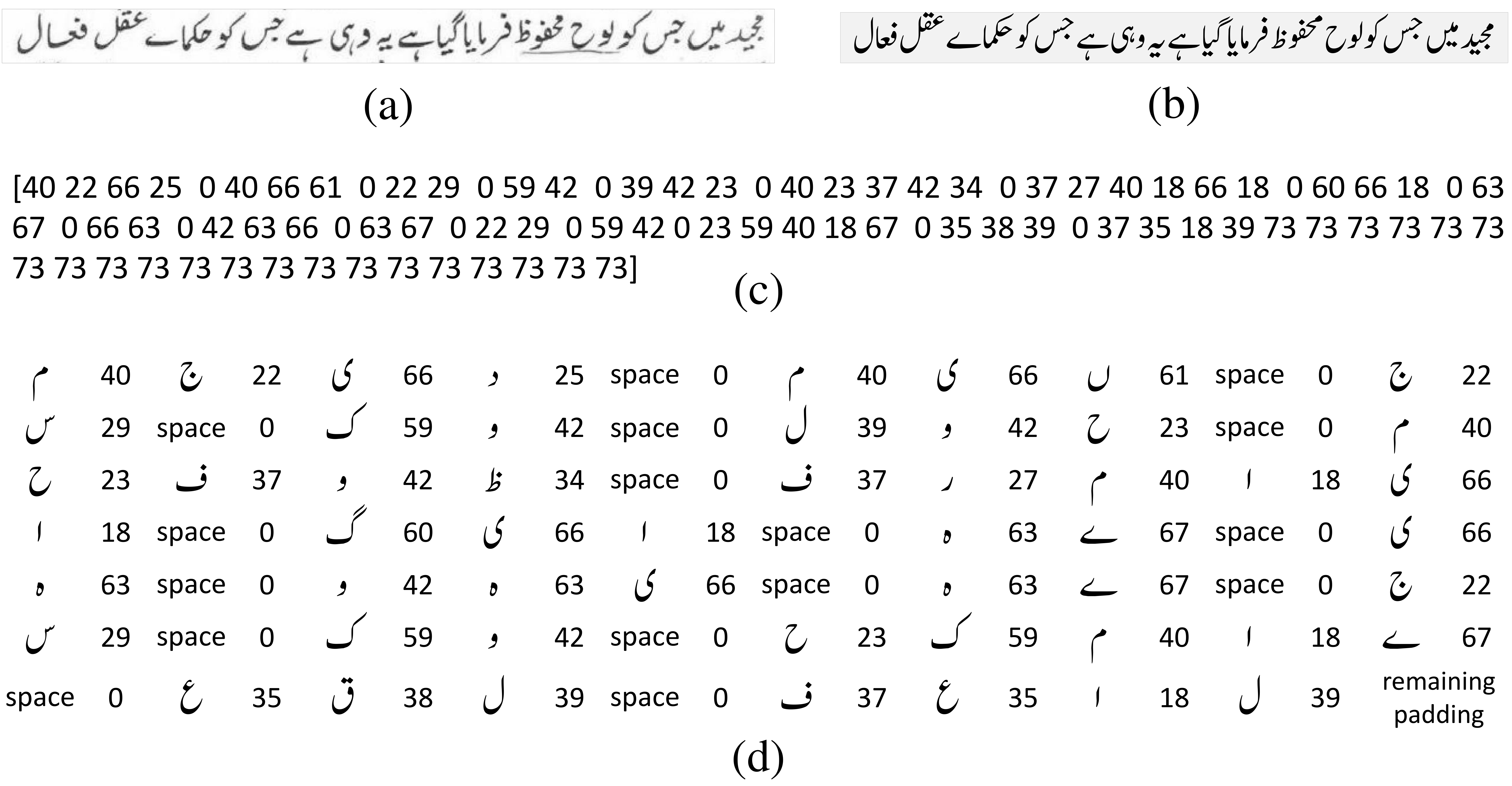}
  \caption{\textbf{(a)} Input Text Line Image, \textbf{(b)} Corresponding Label/ Image Transcription: This is the original text which is a sequence of characters, \textbf{(c)} Preprocessed Label: Each character in the label is replaced by its numeric identifier which creates a sequence of numbers that represents the original text; \textit{actual sequence length is 69, after padding it becomes 90}, \textbf{(d)} Encoding Detail: Showing the unique numeric identifiers assigned to each character in the input sequence (label), i.e. how each character is mapped to a number.}
  \Description{}
  \label{fig:preprocessing-labels}
\end{figure}

An example demonstrating the label preprocessing is depicted in Figure \ref{fig:preprocessing-labels}.

\subsection{Training Process}
A rigorous training phase has been executed, which included the following key steps aimed at optimizing the model performance and ensuring its generalizability.

\subsubsection{Loss Calculation}
The designed models were trained using the CTC loss function, which handles the misalignments between input and target sequences. During training, the loss is computed by considering all possible alignments of the target sequence, which is then propagated back to the network to update the weights and biases. It enabled end-to-end training.

\subsubsection{Hyperparameter Tuning}
Below hyperparameters were tuned carefully, as they significantly impact the model convergence, stable training, and final recognition accuracy.

\begin{itemize}
    \item[$\bullet$] \textbf{Learning Rate:} It is a key hyperparameter that determines the step size taken during model training, affecting convergence speed to optimal weights. Finding an ideal learning rate can be challenging, as a larger rate may skip the optimal solution, while a smaller one may prolong training and often get stuck in local minima. So, various initial learning rates including 0.001, 0.0001, 0.0002, 0.0003, and 0.0004 were examined to find the most suitable one while conducting experiments. Additionally, \textit{learning rate schedulers} were used that dynamically adjusted the learning rate during training.
    \item[$\bullet$] \textbf{Optimizer:} During training it is responsible for updating model parameters i.e. weights and biases, to minimize the loss. It ensures that the model iteratively refines these parameters until convergence. Various optimizers, including Adam, Nadam, Adamax, and RMSprop, were analyzed to identify which one achieves faster convergence and improved performance; \textit{each optimizer employs its own strategy for managing gradient updates and learning rates}.
    \item[$\bullet$] \textbf{Batch Size:} It defines the number of training samples used in each epoch of the training process. Adjusting the batch size can affect the speed and stability of training. Thus, experiments were conducted using various batch sizes including 32, 64, and 128, to access their effects on training dynamics and performance.
\end{itemize}

\subsubsection{Counteract Overfitting}
Overfitting occurs when a model is too complex and fits or specialized on training data, resulting in a lack of generalizability. To counteract the risk of overfitting, the following techniques were adopted:

\begin{itemize}
    \item[$\bullet$] \textbf{Dropout Layers:} These layers played a pivotal role in training by randomly deactivating a fraction of neurons \textit{(dropout rate was 0.3)} during each training iteration. This deliberate dropout of neurons prevented the model from memorizing the training data excessively, thereby enhancing its ability to generalize to unseen data.
    \item[$\bullet$] \textbf{Batch Normalization:} It is achieved by normalizing the input values to each layer within the network. It is a powerful technique that stabilize training, enhance convergence and potentially achieves better generalization \cite{ioffe2015batch}. So, batch normalization layers were incorporated into the CNN component of the designed model architectures.
    \item[$\bullet$] \textbf{Learning Rate Scheduling:} This technique adjusts the learning rate during model training to help the model learn better and avoid overfitting. Hence, the models were trained using schedulers, keeping the learning rate (higher rate for fast learning) constant for some initial epochs and then decaying it exponentially to fine-tune the model.
    \item[$\bullet$] \textbf{Early Stopping:} A regularization technique called early stopping has also been employed to prevent from overfitting and optimizing training efficiency. It involved monitoring the model's performance on validation data and halting the training if there was no improvement over 33 \textit{(the patience value)} consecutive epochs. This ensured that the model stopped training after reaching optimal performance.
\end{itemize}

\subsection{Evaluation Metrics}
\begin{figure}[h!]
  \centering     
  \includegraphics[width=11cm]{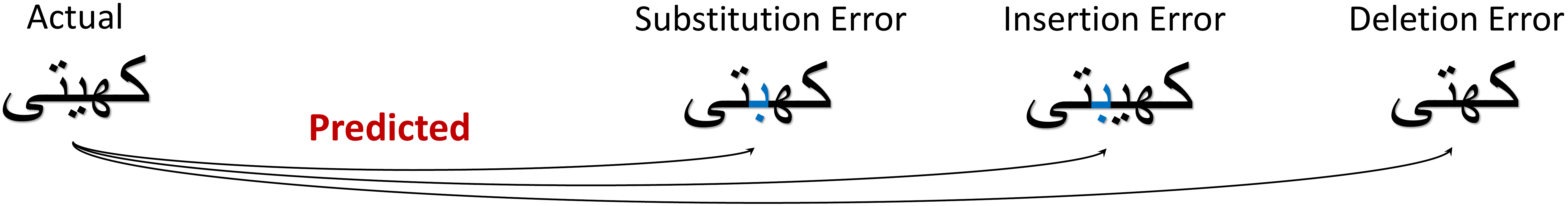}
  \caption{Types of Recognition Errors: \textbf{Substitution} error occurs when a character is incorrectly replaced/misspelled in the predicted text. \textbf{Insertion} error occurs when a character is erroneously included, whereas \textbf{Deletion} errors occur when a character is skipped/omitted in the predicted text.}
  \Description{}
  \label{fig:recognition-error}
\end{figure}

Accuracy is mostly used evaluation metric for accessing the predicted output where `1' indicates matched and `0' denotes no match. However, it does not provide a sufficiently detailed evaluation of the HTR model performance. Therefore, error rates i.e. CER and WER, have been used to gauge the dissimilarity between the predicted text and the actual/reference text. There are three distinct types of recognition errors: Substitution, Insertion, and Deletion error as shown in Figure \ref{fig:recognition-error}. 

The \textbf{Character Error Rate} (CER) which relies on the concept of \textit{Levenshtein distance}\footnote{The \textit{\textbf{Levenshtein Distance}} or \textit{\textbf{Edit Distance}} measures the distance between two strings by calculating the number of edits (insertions, deletions, substitutions) required to transform one string into other.}, is defined by the number of substitutions \textit{(S)}, insertions \textit{(I)}, and deletions \textit{(D)} at character level needed to transform the reference text into the predicted text, divided by the total number of characters \textit{(N)} in the reference text, see Eq. (\ref{eq:cer}). It measures the rate at which characters in the recognized text deviate from the ground truth. By substituting characters with words in Eq. (\ref{eq:cer}), the \textbf{Word Error Rate} (WER) can be computed between the predicted and reference sentences. To convert both error rates into recognition rates, subtract them from 100 \cite{anjum2020attention,gader2022attention-arabic}.

\begin{equation}
    \label{eq:cer}
   CER =  \frac{(S+I+D)}{N} \times 100
\end{equation}

Besides evaluating the CER and WER on validation and test sets, the training and validation loss has also been considered, as it provides insight into how well the model is learning during training and generalizing to unseen data.

\section{Results and Discussion}
\label{sec:results}
This section presents the findings and insightful discussions on the performance and implications of the hybrid models used in this study for UKHR. Let's dive into the details and explore what these findings reveal.

\subsection{Optimizing Models: Hyperparameters Fine-Tuning and Architecture Exploration}
To optimize each of the four models designed for UKHR, we first fine-tuned the hyperparameters including learning rate, batch size, and optimizer by assessing their impact on model’s performance using evaluation metrics. With the fine-tuned hyperparameter settings, we then further examined the multiple architectural variations of each model by testing different configurations to identify the most effective architecture. Let’s see this entire process of finding the optimal settings for the models in detail.

\begin{table*}[ht!]
  \caption{Impact of Hyperparameter Tuning on CNN-BGRU-CTC(1) Model Performance \textemdash Determined Optimal Hyperparameters are: Learning\_Rate (lr)=0.0002, Optimizer (op)=RMSprop, Batch\_Size (bs)=32}
  \label{tbl:bgru-model-hpt}
  \begin{tabularx}{\textwidth}{>{\centering\arraybackslash}p{1.5cm}>{\centering\arraybackslash}p{1.8cm}>{\centering\arraybackslash}p{2cm}>{\centering\arraybackslash}p{1cm}>{\centering\arraybackslash}X>{\centering\arraybackslash}X>{\centering\arraybackslash}X>{\centering\arraybackslash}X>{\centering\arraybackslash}X>{\centering\arraybackslash}X}
    \toprule
     & & & & \multicolumn{2}{c}{Loss(\%)} & \multicolumn{2}{c}{CER(\%)} & \multicolumn{2}{c}{WER(\%)} \\

    \multirow{-2}{1.5cm}{\centering Hyper
    Param}& \multirow{-2}{1.8cm}{\centering Tuning Value}& \multirow{-2}{2cm}{\centering Other Params Values}& \multirow{-2}{1cm}{\centering Epochs}& train & valid & valid & test & valid & test\\
    
    \midrule
    \multirow{5}{1.5cm}{\centering learning rate (lr)} & 0.001 &\multirow{5}{1.8cm}{\centering op=Adam, bs=64} & 113 & 20.5 & 25.4 & 9.8 & 10.0 & 30.3 & 31.3\\
     & 0.0001 &  & 97 & 4.9 & 26.6 & 6.1 & 6.3 & 20.4 & 20.5\\

     & \cellcolor{RubineRed!20}\textbf{0.0002} & & \cellcolor{RubineRed!20}\textbf{70} & \cellcolor{RubineRed!20}\textbf{4.6} & \cellcolor{RubineRed!20}\textbf{22.6} & \cellcolor{RubineRed!20}\textbf{6.3} & \cellcolor{RubineRed!20}\textbf{6.2} & \cellcolor{RubineRed!20}\textbf{20.8} & \cellcolor{RubineRed!20}\textbf{19.9}\\
     & 0.0003 & & 61 & 5.2 & 28.8 & 6.3 & 6.4 & 20.9 & 20.7\\
     & 0.0004 & & 57 & 5.4 & 22.0 & 6.8 & 6.8 & 21.1 & 20.8\\

    \addlinespace

    \multirow{4}{1.5cm}{\centering optimizer (op)} & Adam &\multirow{4}{1.8cm}{\centering lr=0.0002, bs=64} & 70 & 4.6 & 22.6 & 6.3 & 6.2 & 20.8 & 19.9 \\
    & Nadam & & 68 & 4.3 & 21.1 & 6.3 & 6.3 & 19.7 & 19.7 \\
    & \cellcolor{yellow!30}\textbf{RMSprop} & & \cellcolor{yellow!30}\textbf{78} & \cellcolor{yellow!30}\textbf{4.2} & \cellcolor{yellow!30}\textbf{21.0} & \cellcolor{yellow!30}\textbf{6.1} & \cellcolor{yellow!30}\textbf{6.2} & \cellcolor{yellow!30}\textbf{19.8} & \cellcolor{yellow!30}\textbf{19.5} \\
    & Adamax & & 85 & 9.4 & 22.1 & 6.8 & 6.9 & 22.4 & 22.5 \\

    \addlinespace

    \multirow{3}{1.5cm}{\centering batch size (bs)} & \cellcolor{lightblue}\textbf{32} &\multirow{3}{1.8cm}{\centering lr=0.0002, op=RMSprop} & \cellcolor{lightblue}\textbf{59} & \cellcolor{lightblue}\textbf{5.1} & \cellcolor{lightblue}\textbf{21.1} & \cellcolor{lightblue}\textbf{6.1} & \cellcolor{lightblue}\textbf{6.3} & \cellcolor{lightblue}\textbf{19.6} & \cellcolor{lightblue}\textbf{20.2} \\
     & 64 & & 78 & 4.2 & 21.0 & 6.1 & 6.2 & 19.8 & 19.5 \\

    & 128 & & 94 & 4.1 & 21.7 & 6.3 & 6.2 & 19.9 & 19.9 \\
    \bottomrule
    \end{tabularx} 
\end{table*}

To find the optimal hyperparameters for CNN-BGRU-CTC model, we began by fine-tuning the learning rate. In these experiments, the batch size was set to 64, and the Adam optimizer was used as it is the most used optimizer due to its speed and efficiency; \textit{considered a good optimizer as a starting point}. Early stopping with a patience value of 33 was also employed to enhance the effectiveness of the model training process. Table \ref{tbl:bgru-model-hpt} provides insights into the impact of varying learning rates. It can be observed that as the learning rate decreased from 0.001 to 0.0004, there was a noticeable improvement in the model performance, as indicated by lower loss percentages and error rates on both the validation and test sets. Based on the results, the learning rate of 0.0002 stand out as an optimal choice, offering a good balance between training speed and model performance.

Further experiments were carried out using the fixed determined initial learning rate of 0.0002 and a batch size of 64, but with different optimizers. By accessing the different optimizers’ performance given in Table \ref{tbl:bgru-model-hpt}, RMSprop displayed the best results while Adam and Nadam also performed well. However, Adamax yielded less favorable outcomes. Therefore, the RMSprop optimizer was selected for further experiments due to its superior optimization capabilities.

After determining the optimal learning rate (0.0002) and optimizer (RMSprop), the impact of various batch sizes on model performance was investigated. Results presented in Table \ref{tbl:bgru-model-hpt} reveals that a batch size of 32 delivered the best results, indicating a well-balanced training efficiency and model performance. Conversely, batch sizes of 64 and 128 led to slightly less optimal performance, with slightly higher error rates.

\begin{table*}[h!]
  \caption{Performance Comparison among Four Variants of the CNN-BGRU-CTC Model Architecture under Optimized Hyperparameters}
  \label{tbl:bgru-models-architectures}
  \begin{tabularx}{\textwidth}{>{\centering\arraybackslash}p{1.8cm}>{\centering\arraybackslash}p{3cm}>{\centering\arraybackslash}p{1cm}>{\centering\arraybackslash}p{1cm}>{\centering\arraybackslash}X>{\centering\arraybackslash}X>{\centering\arraybackslash}X>{\centering\arraybackslash}X>{\centering\arraybackslash}X>{\centering\arraybackslash}X}
    \toprule
     & & & & \multicolumn{2}{c}{Loss(\%)} & \multicolumn{2}{c}{CER(\%)} & \multicolumn{2}{c}{WER(\%)} \\

    \multirow{-2}{1.8cm}{\centering Model Variant}& \multirow{-2}{3cm}{\centering Configurations} & \multirow{-2}{1cm}{\centering Hyper Params} & \multirow{-2}{1cm}{\centering Epochs}& train & valid & valid & test & valid & test\\
    
    \midrule       
    
    \multirow{3}{1.8cm}{\centering CNN-BGRU-CTC(1)} & 7 Conv2D (64, 128, 256*[3], 512*[2]) and 3 BGRU (512) & \multirow{10}{1cm}{\centering \rotatebox[origin=c]{90}{\parbox{3cm}{\centering \textbf{lr=0.0002 \\ op=RMSprop \\ bs=32}}}} & \multirow{3}{1cm}{\centering 59} & \multirow{3}{1cm}{5.1} & \multirow{3}{1cm}{21.1} & \multirow{3}{1cm}{6.1} & \multirow{3}{1cm}{6.3} & \multirow{3}{1cm}{19.6} & \multirow{3}{1cm}{20.2} \\
    
    \addlinespace
    
    \cellcolor{lightblue}\multirow{3}{2cm}{\centering \textbf{CNN-BGRU-CTC(2)}} & \cellcolor{lightblue}\textbf{7 Conv2D (64, 128, 256*[2], 512*[3]) and 3 BGRU (512)} &  & \cellcolor{lightblue}\multirow{3}{1cm}{\centering \textbf{70}} & \cellcolor{lightblue}\multirow{3}{1cm}{\textbf{3.9}} & \cellcolor{lightblue}\multirow{3}{1cm}{\textbf{21.7}} & \cellcolor{lightblue}\multirow{3}{1cm}{\textbf{5.8}} & \cellcolor{lightblue}\multirow{3}{1cm}{\textbf{5.7}} & \cellcolor{lightblue}\multirow{3}{1cm}{\textbf{18.9}} & \cellcolor{lightblue}\multirow{3}{1cm}{\textbf{18.4}} \\

    \addlinespace
    
    \multirow{3}{1.8cm}{\centering CNN-BGRU-CTC(3)} &
    8 Conv2D (64, 128, 256*[3], 512*[3]) and 3 BGRU (512) & & \multirow{3}{1cm}{\centering 78} & \multirow{3}{1cm}{3.3} & \multirow{3}{1cm}{21.7} & \multirow{3}{1cm}{5.9} & \multirow{3}{1cm}{6.0} & \multirow{3}{1cm}{18.9} & \multirow{3}{1cm}{19.1} \\
    
    \addlinespace
     
    \multirow{3}{1.8cm}{\centering CNN-BGRU-CTC(4)} & 
    7 Conv2D (64, 128, 256*[2], 512*[3]) and 4 BGRU (512) & & \multirow{3}{1cm}{\centering 80} & \multirow{3}{1cm}{4.1} & \multirow{3}{1cm}{19.8} & \multirow{3}{1cm}{5.9} & \multirow{3}{1cm}{5.8} & \multirow{3}{1cm}{18.7} & \multirow{3}{1cm}{18.4} \\
            
    \bottomrule
    \end{tabularx} 
\end{table*}

With these fine-tuned hyperparameter settings (lr=0.0002, op=RMSprop, bs=32), further different CNN-BGRU-CTC model architectures were assessed to select the most optimal one. We employed deeper and more complex architectures, their configurations and achieved results are provided in Table \ref{tbl:bgru-models-architectures}. The numerical identifier \textit{(i.e. (1), (2) etc.)} added to the model name denotes a unique variant within the CNN-BGRU-CTC model architecture. Upon the thorough analysis of these results, it is observed that CNN-BGRU-CTC(2) architecture exhibited the most favorable performance. The CNN-BGRU-CTC(3) and CNN-BGRU-CTC(4) architectures also displayed competitive results, showing high model accuracy but the CNN-BGRU-CTC(1) architecture showed slightly less optimal outcomes. It is observed that the deeper model architectures like CNN-BGRU-CTC(3) and CNN-BGRU-CTC(4) required more training time while the achieved results were almost near to the results of CNN-BGRU-CTC(2) architecture. Ultimately, by incorporating the model’s generalizability, complexity and performance the CNN-BGRU-CTC(2) architecture was selected as the \textit{final CNN-BGRU-CTC model architecture}.

In a nutshell, adopting the same systematic approach as adopted for CNN-BGRU-CTC model, the optimal hyperparameters and model architecture are identified for other three hybrid models: the CNN-LSTM-CTC model, CNN-GRU-CTC model, and the CNN-BLSTM-CTC model. The final achieved results of these models can be seen in Table \ref{tbl:hybrid-models-comparison}.

\subsection{Comparison of CRNN-based Hybrid Models}
The overall performance analysis of the hybrid models developed for UKHR in Table \ref{tbl:hybrid-models-comparison} indicates that the CNN-LSTM-CTC and CNN-GRU-CTC models performed well, but their bidirectional variants outperformed with low loss percentages and error rates. This improved performance can be attributed to the nature of these networks. LSTM and GRU are itself unidirectional, they process the input sequence sequentially from beginning to end and considering only past information, whereas their bidirectional variants i.e. BLSTM and BGRU processes the input sequence in both directions. Therefore, it is concluded that in the context of handwriting recognition, where the input involves image-based sequences, leveraging information from both directions is particularly advantageous. This enables the model to extract richer features and gain better insights, leading to more accurate recognition. Consequently, among the evaluated models, the CNN-BGRU-CTC model stood out as the best-performing model with the lowest CER and WER, and also computationally less expensive than the CNN-BLSTM-CTC model. Hence, CNN, and BGRU along with CTC, was selected as the best architecture for UKHR, so refer to this model as the \textit{UKHR model}. In further discussions, the term UKHR model will be used instead of CNN-BGRU-CTC model.

\begin{table*}[ht!]
  \caption{Performance Comparison of the CRNN-based Hybrid Models for UKHR \textemdash The detailed architecture and configuration of these models can be seen in Figure \ref{fig:ukhr-model} and Table \ref{tbl:ukhr-model-configurtion}, respectively; first understand the architectural diagram (especially its caption), and then see its corresponding configuration table.}
  \label{tbl:hybrid-models-comparison}
  \begin{tabularx}{\textwidth}{>{\centering\arraybackslash}p{2.8cm}>{\centering\arraybackslash}p{3.2cm}>{\centering\arraybackslash}p{1cm}>{\centering\arraybackslash}X>{\centering\arraybackslash}X>{\centering\arraybackslash}X>{\centering\arraybackslash}X>{\centering\arraybackslash}X>{\centering\arraybackslash}X}
    \toprule
     & & & \multicolumn{2}{c}{Loss(\%)} & \multicolumn{2}{c}{CER(\%)} & \multicolumn{2}{c}{WER(\%)} \\

    \multirow{-2}{2.3cm}{\centering Hybrid Model}& \multirow{-2}{3.2cm}{\centering Hyper Params}& \multirow{-2}{1cm}{\centering Epochs}& train & valid & valid & test & valid & test\\
    
    \midrule
    CNN-LSTM-CTC Model & lr=0.0003, op=Nadam, bs=64 & 
    \multirow{2}{*}{77} & 
    \multirow{2}{*}{6.4} & \multirow{2}{*}{21.3} & \multirow{2}{*}{6.2} & \multirow{2}{*}{6.4} & \multirow{2}{*}{20.5} & \multirow{2}{*}{20.6} \\

    \addlinespace
    
    CNN-GRU-CTC Model & lr=0.0003, op=Nadam, bs=64 & \multirow{2}{*}{66} & \multirow{2}{*}{8.9} & \multirow{2}{*}{24.4} & \multirow{2}{*}{6.5} & \multirow{2}{*}{6.7} & \multirow{2}{*}{21.6} & \multirow{2}{*}{21.9} \\

    \addlinespace
    
    CNN-BLSTM-CTC Model & lr=0.0003, op=Adam, bs=64 & \multirow{2}{*}{62} & \multirow{2}{*}{2.8} & \multirow{2}{*}{20.1} & \multirow{2}{*}{6.2} & \multirow{2}{*}{6.2} & \multirow{2}{*}{19.8} & \multirow{2}{*}{19.8} \\
    
    \addlinespace
    \rowcolor{lightblue}
    \textbf{CNN-BGRU-CTC Model} & \textbf{lr=0.0002, op=RMSprop, bs=32} & \multirow{2}{*}{\textbf{70}} & \multirow{2}{*}{\textbf{3.9}} & \multirow{2}{*}{\textbf{21.7}} & \multirow{2}{*}{\textbf{5.8}} & \multirow{2}{*}{\textbf{5.7}} & \multirow{2}{*}{\textbf{18.9}} & \multirow{2}{*}{\textbf{18.4}}\\
    
    \bottomrule
\end{tabularx}
\end{table*}

\begin{table*}[h]
      \caption{Impact of Learning Rate Scheduler on UKHR Model Performance}
      \label{tbl:lr-scheduler-results}
      \begin{tabularx}{\textwidth}{p{1.9cm}>{\centering\arraybackslash}p{3cm}>
      {\centering\arraybackslash}p{1.4cm}>{\centering\arraybackslash}p{1cm}>{\centering\arraybackslash}X>{\centering\arraybackslash}X>{\centering\arraybackslash}X>{\centering\arraybackslash}X>{\centering\arraybackslash}X>{\centering\arraybackslash}X}
        \toprule
         & & & & \multicolumn{2}{c}{Loss(\%)} & \multicolumn{2}{c}{CER(\%)} & \multicolumn{2}{c}{WER(\%)} \\

        \multirow{-2}{1.9cm}{Training Mode} & \multirow{-2}{3cm}{\centering Scheduler Description}& \multirow{-2}{1.4cm}{\centering Hyper Params}& \multirow{-2}{1cm}{\centering Epochs}& train & valid & valid & test & valid & test\\
        
        \midrule
        
            without scheduler & \multirow{2}{*}{\centering ---} & \multirow{7}{1.6cm}{\centering \rotatebox[origin=c]{90}{\parbox{1.6cm}{\centering lr=0.0002 \\ op=RMSprop \\ bs=32}}} & \multirow{2}{*}{70} & \multirow{2}{*}{3.9} & \multirow{2}{*}{21.7} & \multirow{2}{*}{5.8} & \multirow{2}{*}{5.7} & \multirow{2}{*}{18.9} & \multirow{2}{*}{18.4} \\
        \addlinespace
        
        \multirow{2}{1.5cm}{\parbox{1.5cm}{with scheduler}} & lr decay: $e^{-0.1}$ (after 30\textsuperscript{th} epoch) & & \multirow{2}{*}{68} & \multirow{2}{*}{3.3} & \multirow{2}{*}{20.4} & \multirow{2}{*}{5.5} & \multirow{2}{*}{5.6} & \multirow{2}{*}{17.9} & \multirow{2}{*}{18.1} \\

        \addlinespace

        \multirow{2}{1.5cm}{\cellcolor{lightblue}\textbf{\parbox{1.6cm}{with scheduler}}} & \cellcolor{lightblue}\textbf{lr decay: $e^{-0.2}$ (after 25\textsuperscript{th} epoch)} & & \cellcolor{lightblue}\multirow{2}{*}{\textbf{59}} & \cellcolor{lightblue}\multirow{2}{*}{\textbf{3.5}} & \cellcolor{lightblue}\multirow{2}{*}{\textbf{18.4}} & \cellcolor{lightblue}\multirow{2}{*}{\textbf{5.2}} & \cellcolor{lightblue}\multirow{2}{*}{\textbf{5.2}} & \cellcolor{lightblue}\multirow{2}{*}{\textbf{17.2}} & \cellcolor{lightblue}\multirow{2}{*}{\textbf{16.9}} \\
        
        \bottomrule
        \end{tabularx}    
    \end{table*}

\subsection{Impact of Learning Rate Scheduler on UKHR Model Performance}
To further enhance the efficiency of the UKHR model in both training and accuracy, a learning rate scheduling technique has been adopted. The impact of different exponential decaying learning rate schedulers on model's performance is presented in Table \ref{tbl:lr-scheduler-results}. The results indicate that training the model without a scheduler results in higher loss and error rates compared to using a scheduler. Consequently, the exponential decaying learning rate scheduler significantly improved the model's performance. The best results were achieved with a learning rate decay of $e^{-0.2}$ after the 25\textsuperscript{th} epoch. This setup not only reduced the loss and error rates but also required fewer training epochs, indicating a more efficient training process.

\subsection{UKHR Model Output and Analysis of Failure Cases}
\begin{figure*}[h!]
  \centering     
  \includegraphics[width=12.7cm]{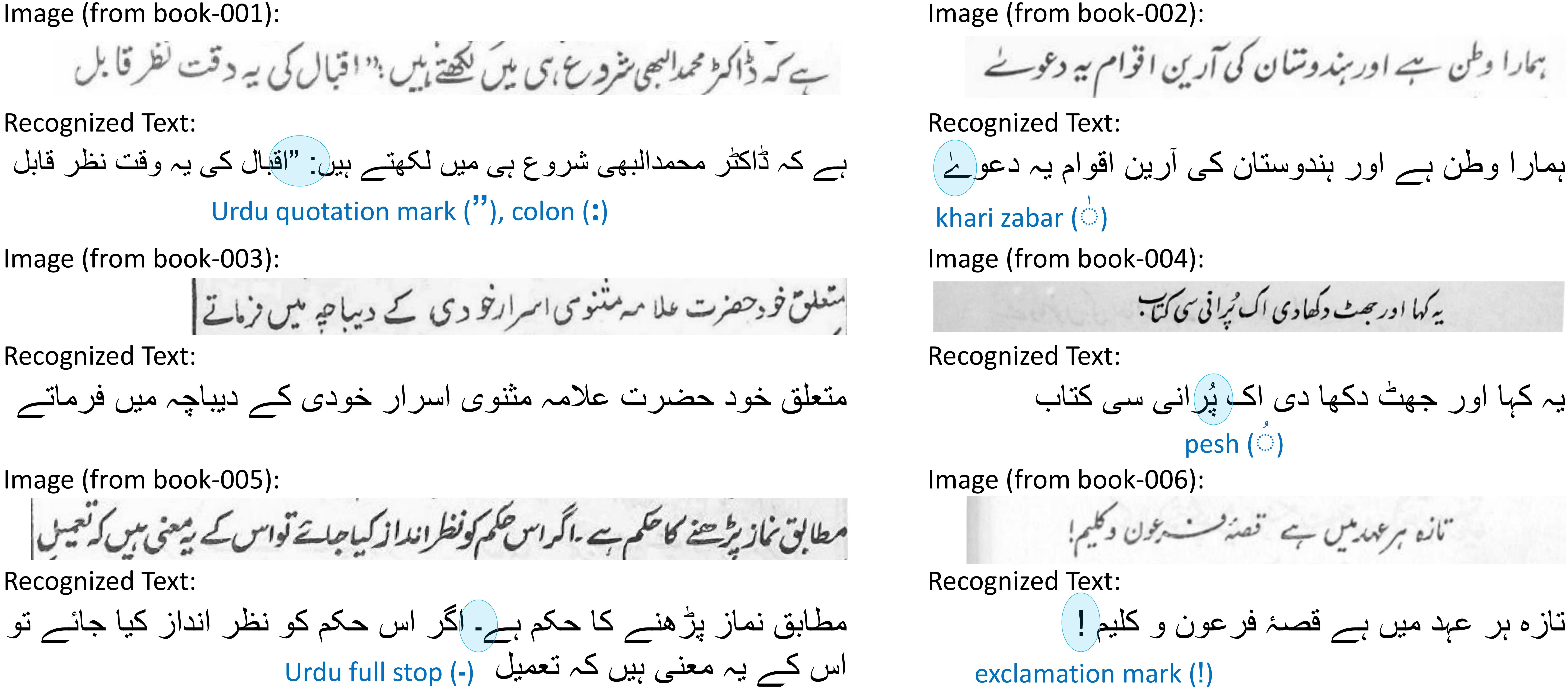}
  \caption{UKHR Model Output -- In Most Cases, It Recognized the Text with 100\% Accuracy}
  \label{fig:ukhr-model-best-cases}
  \Description{}
\end{figure*}

Figure \ref{fig:ukhr-model-best-cases} presents some samples of the input text line images along with predicted texts recognized by the UKHR model. Each sample represents a different katib's handwriting, these images are part of the test set, thus it was unseen data for the model. These recognized texts demonstrate the model's strong performance on unseen data, effectively handling the challenges posed by context sensitivity and overlapping characters in certain contexts. Additionally, it can be seen that the UKHR model accurately recognizes commonly used Urdu punctuation marks, including exclamation marks, quotation marks, full stop, and colon.

\begin{figure*}[h!]
  \centering     
  \includegraphics[width=12.7cm]{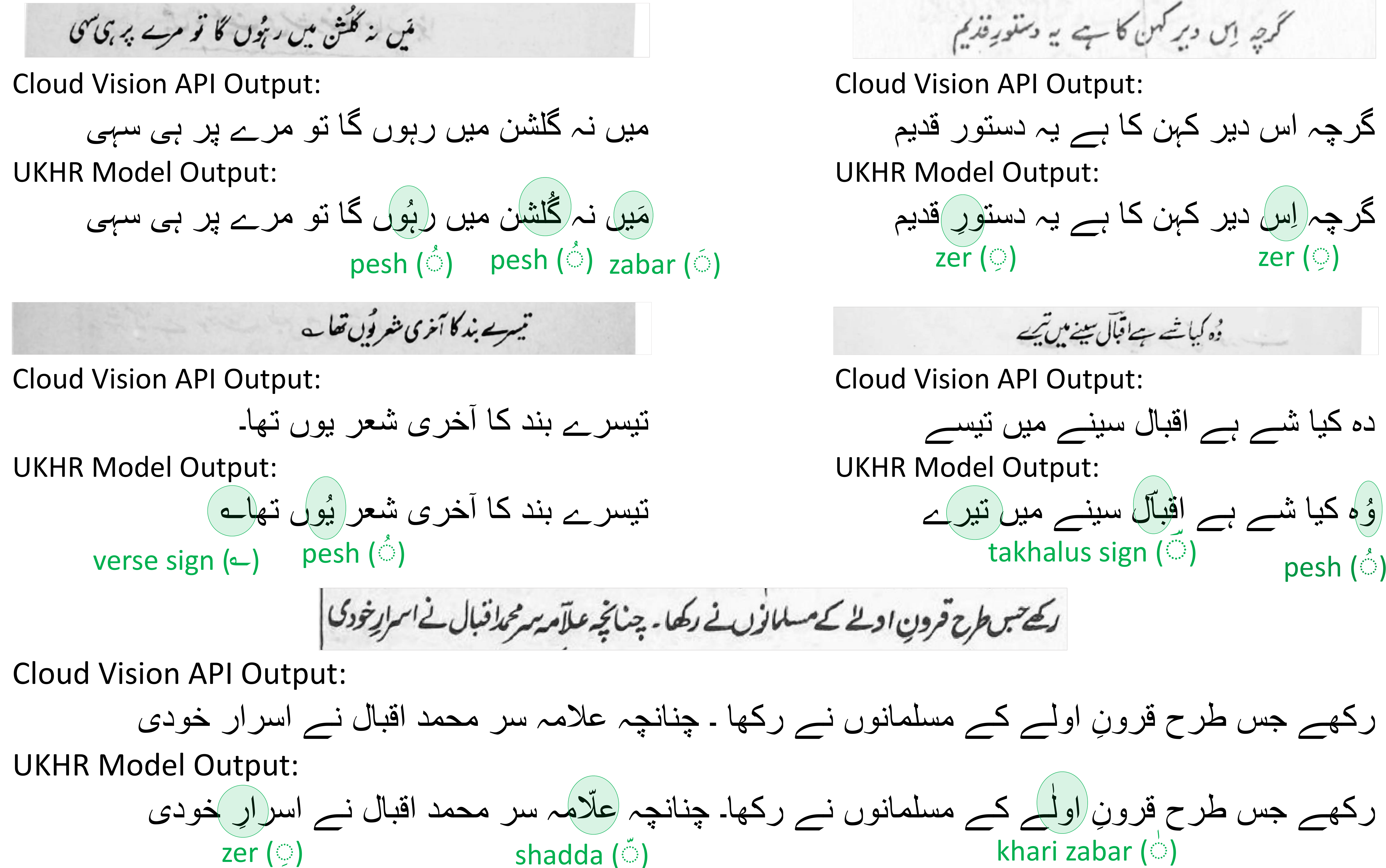}
  \caption{Comparison Between the Output of the Cloud Vision API and the UKHR Model}
  \Description{}
  \label{fig:ukhr-model-vs-google-vision}
\end{figure*}

\begin{figure*}[h!]
  \centering     
  \includegraphics[width=12.7cm]{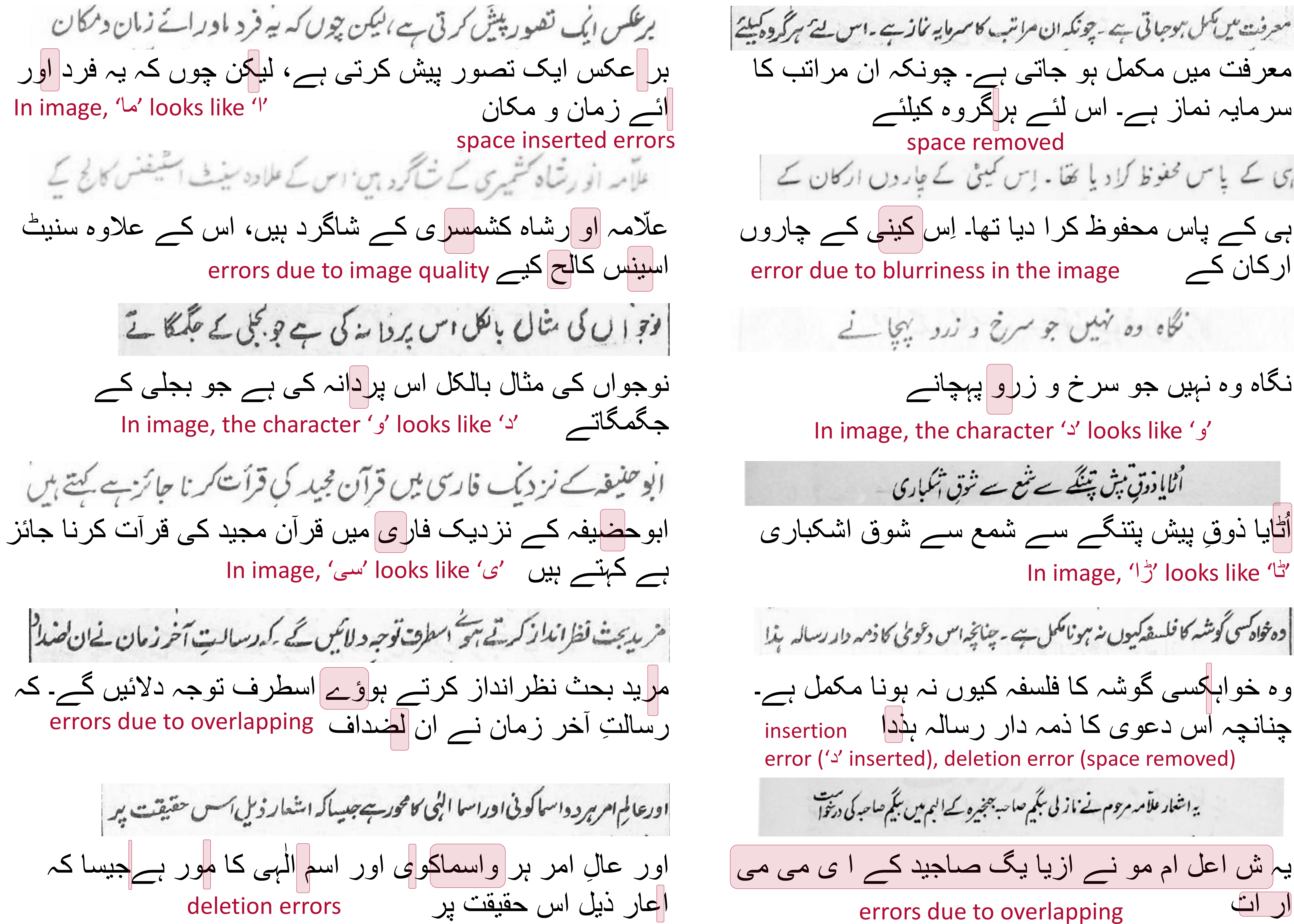}
  \caption{UKHR Model Output --- Failure Case Samples (Recognition Errors are Highlighted with Rectangular Marks)}
  \Description{}
  \label{fig:ukhr-model-failure-cases}
\end{figure*}

As mentioned earlier in the UKHD generation process that we utilized Cloud Vision API for labeling the text line images. Although, it gave good results but also encountered many recognition errors which were then corrected manually. Furthermore, apart from other recognition errors, it was observed that it is not able to recognize diacritics (aerabs). Figure \ref{fig:ukhr-model-vs-google-vision} presents a comparison between the outputs of the cloud vision API and the UKHR model. These predicted texts demonstrate that the UKHR model is able to accurately recognize the mostly used Arabic and Urdu aerabs and symbols such as zer, zabar, pesh, takhalus sign, verse sign etc. whereas google cloud vision is unable to recognize them.

Some examples of failure cases of the UKHR model where it did not recognize the katib’s handwriting correctly are shown in Figure \ref{fig:ukhr-model-failure-cases}. Upon careful examination of these recognized texts, it becomes evident that the model encounters challenges when processing the text with extensive characters overlapping, characters having similar shapes, and also dealing with the low-quality input images, leading to recognition errors.

\section{Conclusion}
\label{sec:conclusion}
This study made notable contributions to advancing the Urdu Handwritten Text Recognition (UHTR). It presents the \textit{Urdu Katib Handwritten Dataset (UKHD)}, a specialized Urdu handwritten text lines dataset curated from the materials written by katibs in historical times. To facilitate the dataset creation process, semi-automatic approaches for segmenting and labeling the text line images have also been introduced which take advantage of existing methods to greatly reduce the time and human effort.

Additionally, the performance of various CRNN-based hybrid models has been evaluated on the primary subset of UKHD, to report the baseline results and best architecture for \textit{Urdu Katib Handwriting Recognition (UKHR)}. These models included CNN-LSTM-CTC, CNN-GRU-CTC, CNN-BLSTM-CTC, and CNN-BGRU-CTC. Among the analyzed models, the CNN-BGRU-CTC model showed the best performance, achieving an average CER and WER of \textbf{5.2\%} and \textbf{16.9\%} on the test set, respectively. We called this model as the \textit{UKHR model}. Although, it performed robustly on unseen data, but it faced challenges in recognizing highly cursive or unconventional handwriting styles, which increased the overall error rates.

In future, the recognition rates could be improved by additional image preprocessing and post-processing of recognized texts. In preprocessing, various image enhancement techniques such as sharpening, contrast enhancement could be utilized to improve the image quality. Because the material used in dataset creation is quite ancient, so images are not in an optimal form. Additionally, future research should explore the integration of transformer architectures with CRNN models for post-processing. We can leverage from pre-trained transformer-based models (e.g. BERT) as their language modeling capability will be helpful in correcting the errors like invalid insertion or deletion of characters or spaces between the words. Apart from that, future research may investigate the development of UHTR systems exclusively using transformer architectures, such as vision transformers and encoder–decoder models. Moreover, experimenting with more diverse and larger datasets as well as exploring advanced optimization techniques could further enhance the model's generalization capabilities.

In conclusion, \textbf{this work is just an initial step} that sets the stage for further exploration and innovation in the domain of UHTR, offering a bridge between tradition and technology. It will support the research community to develop robust recognition systems aimed at preserving Urdu handwritten literature.

\begin{acks}
The authors would like to acknowledge Iqbal Academy Pakistan for providing access to the Iqbal Cyber Library, from which the source books were obtained and used to create the dataset for this research.
\end{acks}

\section*{Data Availability}
The Urdu Katib Handwritten Dataset (UKHD) will be made publicly available upon publication. Prior to its release, researchers interested in using the dataset for academic and non-commercial research purposes may contact the authors.

\bibliographystyle{ACM-Reference-Format}
\bibliography{bibliography}

\appendix

\end{document}